\documentclass{article}

\usepackage{times}

\usepackage{microtype}
\usepackage{graphicx}
\usepackage{subcaption}
\usepackage{booktabs}
\usepackage{fullpage}

\usepackage{hyperref}

\usepackage{xcolor}
\usepackage[vlined, algo2e]{algorithm2e} 
\usepackage{amsmath}
\usepackage{amssymb}
\usepackage{mathtools}
\usepackage{amsthm}
\allowdisplaybreaks
\usepackage{thmtools}
\usepackage{thm-restate}
\usepackage{natbib}
\usepackage[capitalize,noabbrev,nameinlink]{cleveref}
\usepackage{algorithm}
\usepackage{algorithmic}

\newcommand*\samethanks[1][\value{footnote}]{\footnotemark[#1]}

\title{Near-Optimal Regret in Adversarial Kernel Bandits}
\author{
    Yu-Jie Zhang\thanks{Equal contribution.}\\
    University of Washington\\
    \texttt{yujiez7@cs.washington.edu}
    \and
    Hao Qiu\samethanks\\
    National University of Singapore\\
    \texttt{qiuhaosai@gmail.com}
    \and
    Jonathan Scarlett\\
    National University of Singapore \\
    \texttt{scarlett@comp.nus.edu.sg} \\
    \and
    Kevin Jamieson\\
    University of Washington \\
    \texttt{jamieson@cs.washington.edu} \\
}

\usepackage{amsmath}
\usepackage{amssymb}
\usepackage{mathtools}
\usepackage{amsthm}
\usepackage{dsfont}
\usepackage{lipsum}
\usepackage{bm,bbm}
\usepackage{bbding}
\usepackage{pifont}
\usepackage{mathrsfs}
\usepackage{enumitem}
\usepackage{multirow}
\usepackage{graphicx}
\usepackage[font=small,labelfont=bf]{caption}
\usepackage{subfloat}
\usepackage{float}
\usepackage{algorithm,algorithmic}
\usepackage{wrapfig}
\usepackage{booktabs}
\usepackage{verbatim}
\usepackage{setspace}
\usepackage{tcolorbox}
\usepackage{nicefrac}
\usepackage{tikz}
\usepackage{colortbl}
\usepackage{array} % For better alignment
\usepackage{tcolorbox}

\def \E {\mathbb{E}}
\def \x {\mathbf{x}}
\def \y {\mathbf{y}}

\def \z {\mathbf{z}}
\def \u {\mathbf{u}}
\def \H {\mathcal{H}}
\def \w {\mathbf{w}}
\def \O {\mathcal{O}}
\def \R {\mathbb{R}}

\def \F {\mathcal{F}}

\def \v {\mathbf{v}}

\def \F {\mathcal{F}}

\def \u {\mathbf{u}}

\def \X {\mathcal{X}}

\DeclareMathOperator*{\argmax}{arg\,max}
\DeclareMathOperator*{\argmin}{arg\,min}
\renewcommand{\tilde}{\widetilde}
\renewcommand{\hat}{\widehat}

\usepackage{mathtools}
\usepackage{appendix}
\let\norm\undefined % <-- "Undefine" \norm
\DeclarePairedDelimiter\norm{\lVert}{\rVert}

\newcommand\inner[2]{\langle #1, #2 \rangle}
\newcommand{\remarkend}{\hfill\ding{117}}
\theoremstyle{definition}
\newtheorem{myAssum}{Assumption}

\newtheorem{myDef}{Definition}

\newtheorem{myRemark}{Remark}

\renewcommand{\tilde}{\widetilde}
\renewcommand{\hat}{\widehat}

\def \E {\mathbb{E}}

\def \H {\mathcal{H}}

\def \R {\mathbb{R}}

\def \X {\mathcal{X}}

\def \x {\mathbf{x}}
\def \y {\mathbf{y}}

\def \epsilon {\varepsilon}

\definecolor{DSgray}{cmyk}{0,1,0,0}

\usepackage[textsize=tiny]{todonotes}

\usepackage{prettyref}
\newcommand{\pref}[1]{\prettyref{#1}}

\newcommand{\savehyperref}[2]{\texorpdfstring{\hyperref[#1]{#2}}{#2}}
\newrefformat{eq}{\savehyperref{#1}{Eq.~(\ref*{#1})}}
\newrefformat{eqn}{\savehyperref{#1}{Equation~\ref*{#1}}}
\newrefformat{lem}{\savehyperref{#1}{Lemma~\ref*{#1}}}
\newrefformat{lemma}{\savehyperref{#1}{Lemma~\ref*{#1}}}
\newrefformat{def}{\savehyperref{#1}{Definition~\ref*{#1}}}
\newrefformat{line}{\savehyperref{#1}{Line~\ref*{#1}}}
\newrefformat{thm}{\savehyperref{#1}{Theorem~\ref*{#1}}}
\newrefformat{corr}{\savehyperref{#1}{Corollary~\ref*{#1}}}
\newrefformat{cor}{\savehyperref{#1}{Corollary~\ref*{#1}}}
\newrefformat{sec}{\savehyperref{#1}{Section~\ref*{#1}}}
\newrefformat{app}{\savehyperref{#1}{Appendix~\ref*{#1}}}
\newrefformat{appendix}{\savehyperref{#1}{Appendix~\ref*{#1}}}
\newrefformat{assum}{\savehyperref{#1}{Assumption~\ref*{#1}}}
\newrefformat{ex}{\savehyperref{#1}{Example~\ref*{#1}}}
\newrefformat{fig}{\savehyperref{#1}{Figure~\ref*{#1}}}
\newrefformat{alg}{\savehyperref{#1}{Algorithm~\ref*{#1}}}
\newrefformat{rem}{\savehyperref{#1}{Remark~\ref*{#1}}}
\newrefformat{conj}{\savehyperref{#1}{Conjecture~\ref*{#1}}}
\newrefformat{prop}{\savehyperref{#1}{Proposition~\ref*{#1}}}
\newrefformat{proto}{\savehyperref{#1}{Protocol~\ref*{#1}}}
\newrefformat{prob}{\savehyperref{#1}{Problem~\ref*{#1}}}
\newrefformat{claim}{\savehyperref{#1}{Claim~\ref*{#1}}}
\newrefformat{que}{\savehyperref{#1}{Question~\ref*{#1}}}
\newrefformat{op}{\savehyperref{#1}{Open Problem~\ref*{#1}}}
\newrefformat{fn}{\savehyperref{#1}{Footnote~\ref*{#1}}}
\allowdisplaybreaks[3]

\newcommand{\ke}{\mathcal{K}}          % kernel function K(x,y)
\newcommand{\Reg}{\mathcal{R}}         % regret R_n
\newcommand{\wht}{\hat{\w}_t}           % biased loss estimator
\newcommand{\SigLam}{\Sigma_t^{\lambda}}
\usepackage{thmtools}
\usepackage{thm-restate} % Provides the restatable 

\newcommand{\Tr}{\mathrm{Tr}}
\begin{document}
\maketitle

\begin{abstract}
We study the adversarial kernel bandit problem, in which the loss at each round is induced by an arbitrary bounded element of a reproducing kernel Hilbert space (RKHS). 
We propose an exponential-weights algorithm built on a regularized importance-weighted loss estimator, together with an explicit correction term that cancels the bias introduced by the regularization. 
Our main result bounds the regret by $\widetilde{\mathcal{O}}\big(\sqrt{T\, d_*(\lambda)\,\log|\X|}\big)$, where $d_*(\lambda)$ is a widely-adopted notion of effective dimension that captures the complexity of the kernel. 
Up to logarithmic factors, this matches the known rate achieved in the related stochastic kernel bandit problem. 
A notable application is the Mat\'ern$(\nu,d)$ kernel with smoothness parameter $\nu$ on $\R^d$, for which our bound specializes to $\widetilde{\mathcal{O}}\big(T^{(\nu+d)/(2\nu+d)}\big)$, improving over the best-known prior rate of \citet{chatterji2019online} while simultaneously removing the rank-one adversary assumption required by their analysis. 
Moreover, this rate is the same as the known optimal rate for stochastic kernel bandits, and also matches a lower bound from concurrent work up to a $\log T$ factor.
\end{abstract}

\section{Introduction}
Online decision-making under bandit feedback (e.g., see \cite{cesa2006prediction, hazan2016introduction, lattimore2020bandit}) is a fundamental problem in machine learning, with applications ranging from recommendation systems~\citep{li2010contextual} to experimental design~\citep{srinivas2010gp}. In the classical adversarial bandit setting, each arm has an arbitrary loss sequence, a learner repeatedly selects actions and observes only the loss incurred by the chosen action, with the goal of competing against the best fixed action in hindsight. A well-studied extension of this framework is adversarial linear bandits~\citep{awerbuch2004adaptive,conf/nips/DaniHK07,bubeck2012towards}, where the loss of an action at each time step is assumed to be a linear function of its features. While this linear structure enables efficient learning and strong regret guarantees, it is often too restrictive to capture complex, nonlinear relationships present in modern applications.

To address this limitation, we study the \emph{adversarial kernel bandits} problem~\citep{chatterji2019online, takemori2021approximation}, in which losses are modeled as functions in a reproducing kernel Hilbert space (RKHS). This framework strictly generalizes linear bandits by allowing nonlinear losses with a potentially infinite-dimensional feature representation.  It serves as an adversarially time-varying counterpart to the well-studied stochastic kernel bandit problem, which is typically viewed through the lens of Bayesian optimization via Gaussian process methods \citep{garnett_bayesoptbook_2023}.  
The added expressivity of RKHS functions (compared to linear functions) can be highly valuable in diverse applications.  This is evidenced by the extensive applications of the stochastic formulation, such as hyperparameter tuning \cite{chen2018bayesian}, molecule design \cite{korovina2020chembo}, environmental monitoring \cite{marchant2012bayesian}, and many more.  The adversarial formulation is of interest in scenarios where the underlying function may be subject to unknown dynamic behavior, changepoints, and so on, and also has a strong theoretical precedent based on the extensive work on adversarial (non-kernel) bandits. 

The additional expressiveness of this setup introduces significant challenges: under bandit feedback, the learner observes only a single scalar loss per round, making it difficult to estimate loss functions whose underlying structure may be infinite-dimensional. Despite its practical and theoretical importance, existing regret guarantees for adversarial kernel bandits remain incomplete for some of the most commonly used kernels. The closest prior work, \citet{chatterji2019online}, considers kernels whose Mercer eigenvalues exhibit either exponential or polynomial decay. In the former case, which includes the squared-exponential (SE) kernel, they obtain a near-optimal $\widetilde{\mathcal{O}}(\sqrt{T})$ rate. Under polynomial eigendecay, however, their bound is no longer tight. This looseness is particularly evident for the Mat\'ern$(\nu,d)$ kernel, a standard choice in Gaussian process modeling and Bayesian optimization. For this kernel, their rate becomes $\widetilde{\mathcal{O}}(T^{(2\nu+d)/(4\nu)})$, which is far from the $\widetilde{\mathcal{O}}(T^{(\nu+d)/(2\nu+d)})$ rate known to be optimal in the analogous stochastic kernel bandit problem. In fact, it fails to be sublinear in $T$ when the smoothness $\nu$ is small relative to the ambient dimension $d$, namely $\nu < d/2$. This gap motivates the central question of this paper:
\begin{quote}
    \centering
\emph{Can one design an adversarial kernel bandit algorithm whose regret \\ matches the optimal stochastic kernel bandit rate (within log factors)?}
\end{quote}

\paragraph{Contributions.} In this paper, we make substantial progress on this question, including an affirmative answer for the Mat\'ern kernel. We propose a simple and principled algorithm for adversarial kernel bandits that combines exponential weights with a regularized importance-weighted loss estimator in the RKHS. The regularization prevents the estimator from becoming unbounded in infinite-dimensional feature spaces, while an explicit correction term compensates for the bias introduced by regularization. This correction is central to the analysis: it cancels the comparator-side bias and leaves only a learner-side residual that can be controlled through a well-known notion of \emph{effective dimension} of the kernel.

Our main result is a regret bound of the form $\widetilde{\mathcal{O}}(\sqrt{T\, d_*(\lambda)\,\log|\X|})$, where $d_*(\lambda)$ is the effective dimension of the kernel and $|\X|$ is the size of the action set. Up to logarithmic factors, this matches the rate achievable in the analogous stochastic kernel bandit problem when expressed via the maximum information gain (as defined in \citep{srinivas2010gp} among many others).  In particular, when specialized to the Mat\'ern$(\nu,d)$ kernel, our bound yields the regret $\widetilde{\mathcal{O}}(T^{(\nu+d)/(2\nu+d)})$, improving over the best-known prior rate $\widetilde{\mathcal{O}}(T^{(2\nu+d)/(4\nu)})$ of \cite{chatterji2019online} while simultaneously removing their restrictive rank-one adversary assumption. This rate further matches the lower bound appearing in a concurrent work~\citep{iwazaki2026nearly} up to logarithmic factors in $T$. For the SE kernel, our bound recovers the near-optimal $\widetilde{\mathcal{O}}(\sqrt{T})$ rate already established by \cite{chatterji2019online}, while again removing the rank-one assumption. Our bound also yields corollaries under suitably-defined polynomial and exponential eigendecay conditions on the kernel spectrum: Under polynomial eigenvalue decay with exponent $\beta>1$, our algorithm achieves
regret
$
    \widetilde{O}\big(T^{(\beta+1)/(2\beta)}\big),
$
and under exponential eigenvalue decay, the regret improves to
$
    \widetilde{O}(\sqrt{T}).
$ 
Although the polynomial-decay rate has the same expression in the exponent as a lower bound from \cite{chatterji2019online}, the two results are proved under different settings and are not directly comparable; see \pref{app:discussion_1} for details.

\paragraph{Concurrent work.} In independent concurrent work, \cite{iwazaki2026nearly} closes the same gap through an approach that is algorithmically and analytically close to ours. Both works use exponential weights with a regularized importance-weighted loss estimator, together with an additive correction term that cancels the bias introduced by the regularization.  There are some minor differences in the problem setup and complexity measures used, but these are essentially inconsequential; see \pref{app:discussion_2} for the details.  

Beyond this shared core, \cite{iwazaki2026nearly} additionally provides a complete analysis of a Nystr\"om-approximation variant, which we discuss only briefly as a computational simplification, and establishes algorithm-independent lower bounds for the SE and Mat\'ern kernels. % See more discussion and comparison in \pref{app:discussion_2}.

\section{Related Works} \label{sec:related}
\paragraph{Stochastic kernel bandits.} There have been extensive studies on the stochastic kernel bandit problem, in which the underlying function is fixed and the observations are corrupted by noise.  Here we focus only on the most fundamental and/or relevant works.  Prominent early works studied UCB-style algorithms \cite{srinivas2010gp,chowdhury2017kernelized} and established regret upper bounds of $\tilde{O}(\gamma_T \sqrt{T})$, where $\gamma_T$ is an \emph{information gain} quantity capturing the kernel complexity, and is closely related to the \emph{effective dimension} (e.g., see \cite[Prop.~5] {calandriello2019gaussian}).  This was subsequently improved to $\tilde{O}(\sqrt{T \gamma_T})$ via the SupKernelUCB algorithm \cite{valko2013finite} and subsequently by several other methods \cite{li2022gaussian,camilleri21a,salgia2021domain}.  For the widely-adopted squared exponential (SE) and Mat\'ern kernels, tight upper bounds on $\gamma_T$ were derived in \cite{vakili2021information}, and these bounds in turn establish that the $\tilde{O}(\sqrt{T \gamma_T})$ regret upper bound is tight (to within logarithmic factors) in view of the matching lower bounds proved in \cite{scarlett2017lower,cai2021lower}.  In particular, under polynomial eigenvalue decay with exponent $\beta>1$, the information gain scales as $\widetilde{\mathcal{O}}(T^{1/\beta})$, yielding a regret bound of order $\widetilde{\mathcal{O}}(T^{\frac{\beta+1}{2\beta}})$, whereas for exponential eigenvalue decay the regret becomes $\widetilde{\mathcal{O}}(\sqrt T)$. 
Various works have also studied non-stationary stochastic kernel bandits (see \cite{hong2023optimization,cai2025lower,iwazaki2025near} and the references therein), but their setup is fundamentally different from adversarial kernel bandits, with results from the two settings being incomparable. Specifically, they seek to minimize \emph{dynamic regret} (i.e., track the maximum at every time step) and they crucially place explicit bounds on how much the function can vary. 

\paragraph{Adversarial linear bandits.} 
The adversarial linear bandit problem was introduced to the online learning literature by \citet{awerbuch2004adaptive}. In this setting, a learner repeatedly selects an action from a decision set and incurs a loss that is linear in the chosen action, while the loss vectors may be chosen adversarially. A wide collection of subsequent work has developed efficient algorithms for this problem. In particular, early results established polynomial-time algorithms for compact convex action sets with regret scaling as $\widetilde{O}(d\sqrt{T})$, where $d$ denotes the dimension of the linear bandit problem, using variants of exponential weights combined with additional exploration \citep{conf/nips/DaniHK07,cesa2012combinatorial,bubeck2012towards}.  \cite{conf/colt/AbernethyHR08} proposed the first computationally efficient algorithm that achieves $\widetilde{O}(d^{3/2}\sqrt{T})$ regret using the Following-the-Regularized-Leader framework. 

\paragraph{Adversarial kernel bandits.} The closest prior work on adversarial kernel bandits is \citet{chatterji2019online}, who study a closely related setting under a rank-one adversary. Their guarantees crucially depend on the eigenvalue decay of the kernel. When the Mercer eigenvalues decay exponentially, they obtain the tight $\widetilde{\O}(\sqrt{T})$ regret. However, the situation is less satisfactory under \emph{polynomial} eigendecay. Under polynomial eigendecay with exponent $\beta>1$, the $T^{\beta/(2(\beta-1))}$ dependence in their upper bound does not match the $T^{(\beta+1)/(2\beta)}$ dependence in their lower bound (see also \pref{app:discussion} for some important caveats regarding their lower bound).  For the Mat\'ern$(\nu,d)$ kernel, the upper bound becomes $\widetilde{\O}(T^{(2\nu+d)/(4\nu)})$, which falls short of the optimal stochastic rate $\widetilde{\O}(T^{(\nu+d)/(2\nu+d)})$, and is only sublinear when $d < 2\nu$. A subsequent work~\citep{takemori2021approximation} removed the rank-one adversary assumption by reducing the problem to a misspecified linear bandit via approximation theory. Their algorithm thus extends to more general adversaries, but the resulting $\widetilde{\O}(T^{(\nu+d)/(2\nu)})$ rate is even further from optimality for the Mat\'ern kernel.

Our effective-dimension bound improves on the upper bounds of \citet{chatterji2019online} in two complementary ways. Applied directly to specific kernels, it recovers the same $\widetilde{\O}(\sqrt T)$ rate on the SE kernel, and yields an improved $\widetilde{\O}(T^{(\nu+d)/(2\nu+d)})$ rate on the Mat\'ern$(\nu,d)$ kernel, sharpening their $\widetilde{\O}(T^{(2\nu+d)/(4\nu)})$ bound.  Moreover, under a strengthened version of polynomial eigendecay that holds uniformly over all distributions on $\X$, our bound improves their general rate from $\widetilde{\O}(T^{\beta/(2(\beta-1))})$ to $\widetilde{\O}(T^{(\beta+1)/(2\beta)})$.  We re-iterate that, as we detail in \pref{app:discussion}, their lower bound is not directly comparable to our upper bound despite both having $T^{(\beta+1)/(2\beta)}$ dependence.

\paragraph{Adversarial contextual kernel bandits.} Another relevant work is \cite{neu2024go}, who study an adversarial kernelized \emph{contextual} bandit with $|\X|$ actions and stochastic contexts, where the kernel is defined over the context space and a separate RKHS element is associated with each action. Although their setup is not the same as ours,\footnote{The main difference is that their setup uses kernel modeling for generalizing across contexts, not across actions.} their algorithm can potentially be applied here as well, and it attains the optimal $T$-exponent under polynomial eigendecay. However, because it is built on Follow-the-Regularized-Leader with a \emph{log-barrier} regularizer, it incurs a polynomial dependence on $|\X|$ rather than the $\log|\X|$ dependence afforded by our exponential weights approach.  Thus, there is an \emph{exponentially large gap} between the two dependencies, and this is compounded by the fact that a standard covering argument of a continuous domain suggests choosing $|\X| = T^{d/2}$ or similar \cite{janz2020bandit,li2022gaussian}, making the higher $|\X|$ dependence prohibitive.  We defer a more detailed discussion to Section~\ref{subsec:exp2-challenges}.

\section{Problem Setup}
\label{sec:problem_setup}

This section introduces the problem setup for adversarial kernel bandits, a generalization of adversarial linear bandits in which the loss functions belong to a reproducing kernel Hilbert space (RKHS). We begin by stating the kernel loss model and assumptions, and then present the learning protocol and define the notion of regret.

\paragraph{Kernels and RKHS.} Let $\X \subset \R^d$ be an action space.  We will focus on the case that $\X$ is finite (possibly very large), but our main result extends easily to continuous domains via a standard covering argument, e.g., for the domain $[0,1]^d$, taking a grid of size $T^{d/2}$ leads to negligible discretization error under mild assumptions on the kernel \cite{janz2020bandit,li2022gaussian}.

We are given a symmetric positive definite kernel $\ke:\X\times\X\to\R$, and denote by $\H$ its associated reproducing kernel Hilbert space (RKHS). The RKHS $\H$ is a Hilbert space of functions $f:\X\to\R$, equipped with inner product $\inner{\cdot}{\cdot}_{\H}$ and norm $\norm{\cdot}_{\H}$. It satisfies the reproducing property: for every $f\in\H$ and $\x\in\X$, $f(\x)=\inner{f}{\ke(\cdot,\x)}_{\H}.$ The kernel also induces a feature map $\Phi:\X\to\H$ such that $\ke(\x,\y)=\inner{\Phi(\x)}{\Phi(\y)}_{\H}$ for all $\x,\y\in\X$. The feature vector $\Phi(\x)$ may be infinite-dimensional. When $\ke$ admits a Mercer decomposition under a reference measure, an explicit feature map can be constructed by setting $\Phi_j(\x)=\sqrt{\mu_j}\phi_j(\x)$ for each $j\ge1$, where $\Phi_j(\x)$ denotes the $j$-th coordinate of $\Phi(\x)$, and $\{\mu_j\}_{j\ge1}$ and $\{\phi_j\}_{j\ge1}$ are the eigenvalues and eigenfunctions of the associated kernel integral operator. 
We refer the reader to~\cite[Section 2.1]{chatterji2019online} and \cite{kanagawa2018gaussian} for further background on RKHS functions.

We use the following operator notation on $\H$. For $\v,\w\in\H$, we use the matrix-style notation $\v\w^\top$ to denote the rank-one linear operator on $\H$ defined by $(\v\w^\top)\u=\v\,\inner{\w}{\u}_{\H}$ for all $\u\in\H$. The identity operator on $\H$ is denoted by $I_\H$. For a bounded, self-adjoint, positive semidefinite operator $A:\H\to\H$ and any $\w\in\H$, we define $\norm{\w}_{A}:=\sqrt{\inner{\w}{A\w}_{\H}}$,
which is a seminorm in general and a norm when $A$ is positive definite. We write $A\preceq B$ if $B-A$ is positive semidefinite.

\paragraph{Learning protocol. } Throughout the paper, we use $\mathbb{Z}_{+}$ to denote the set of positive integers and $[m]$ to denote the set $\{1,2, \ldots, m\}$ for $m \in \mathbb{Z}_{+}$. At each round $t \in [T]$,  the learner selects an action $\x_t \in \X$ while the environment simultaneously selects a vector $\w_t \in \H$. The learner then incurs the loss of its decision $\ell_t(\x_t) = \inner{\Phi(\x_t)}{\w_t}_{\H}.$  Afterward, the learner observes only this scalar loss $\ell_t(\x_t)$ and then proceeds to the next round. We consider an oblivious adversary, in which the sequence $\{\w_t\}_{t=1}^T$ is fixed before the interaction begins and does not depend on the learner's realized actions or internal randomness. The goal of the learner is to minimize the expected regret:
\begin{equation*}
    \Reg_T
    \;:=\;
    \E\!\left[\sum_{t=1}^T \inner{\Phi(\x_t)}{\w_t}_{\H}
    \;-\;
     \sum_{t=1}^T \inner{\Phi(\x_*)}{\w_t}_{\H}\right],
\end{equation*}
where $\x_* = \mathop{\arg\min}_{\x\in\X} \sum_{t=1}^T \inner{\Phi(\x)}{\w_t}_{\H}$ denotes the best fixed action in hindsight and the expectation is over the randomness in the algorithm. We also impose the following standard regularity assumptions, which are common in the literature \citep{chatterji2019online, li2022gaussian}.

\begin{myAssum}[Bounded Kernel and Adversary]
\label{assum:bounded-kernel}
There exists a constant $B>0$ such that $\norm{\w_t}_{\H} \le B$ for all $t \in [T]$. The kernel is bounded by $\ke(\x,\x) \leq 1$ for any $\x\in\X$.\footnote{This assumption easily generalizes to any upper bound of the form $\ke(\x,\x) \leq G^2$ for a fixed constant $G$, but we focus on $\ke(\x,\x) \le 1$ to reduce notation and since this normalization is very standard in the stochastic kernel bandit literature (e.g., see \cite{srinivas2010gp}).}
\end{myAssum}

\begin{figure}[!t]
	\begin{centering}
		\includegraphics[width=0.8\textwidth]{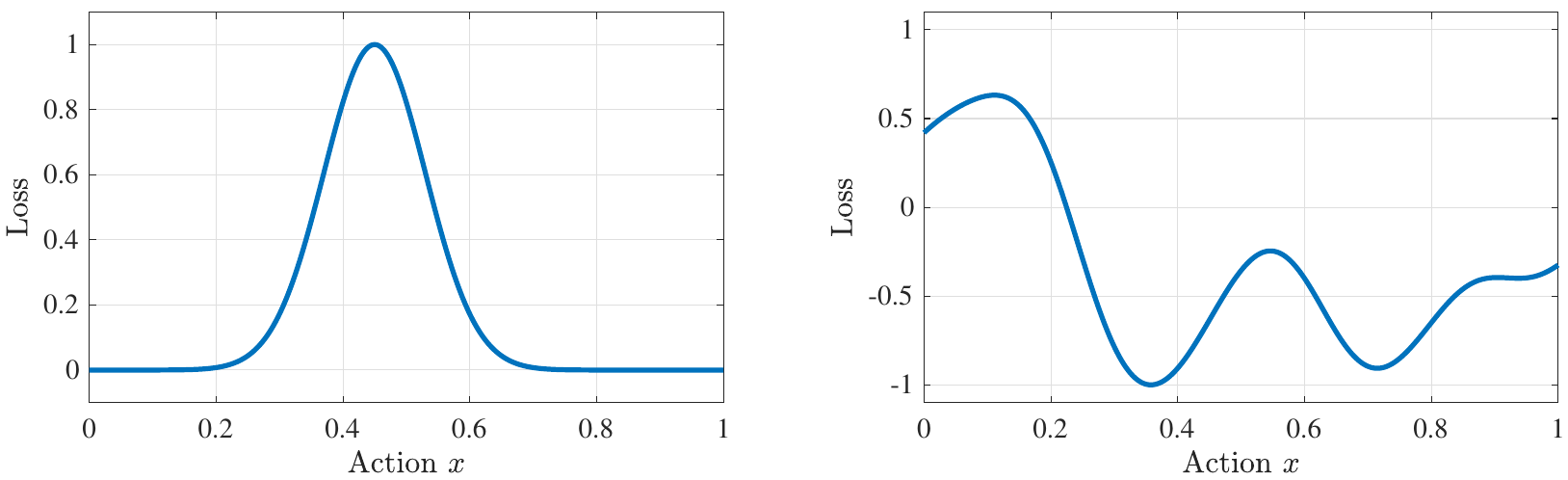} \par
	\end{centering}
	
	\caption{Example of a rank-one function (Left) and a general RKHS-norm bounded function (Right) under the squared exponential kernel, which has exponential eigenvalue decay. \label{fig:rank_one}}
\end{figure}

\begin{myRemark}
\label{rem:rank-one}
We note that the adversarial kernel bandit model of~\citet{chatterji2019online} imposes a \emph{rank-one} restriction on the adversary, which requires $\w_t=\Phi(\x_t)$ for some $\x_t\in\X$.  This is quite restrictive, and to our knowledge it has not been considered in any works on stochastic kernel bandits.  Our framework removes this restriction, allowing $\w_t$ to be an arbitrary element of the RKHS~$\H$ subject only to a bounded norm condition.  See Figure \ref{fig:rank_one} for an illustration.  The rank-one restriction in~\citet{chatterji2019online} is tied to their projection-based loss estimator, whereas our use of a regularized loss estimator frees us from this requirement.
\end{myRemark}

\section{Exponential Weights with Regularized Estimator}
\label{sec:method-exp2}

In this section, we develop a regularized variant of Exp2 for adversarial kernel bandits and derive its regret in terms of the effective dimension. Up to logarithmic factors, this matches the rate achievable in the analogous stochastic kernel bandit problem, and specializes to near-optimal rates for commonly used kernels such as the SE and Mat\'ern kernels. We first step through the challenges that arise when extending Exp2 from linear to kernel losses and explain why existing approaches fall short, then present the algorithm and its regret guarantee.

\subsection{Exp2 and the Kernel Obstacle}
\label{subsec:exp2-challenges}

A central challenge in moving from adversarial linear bandits to kernel bandits is that the underlying parameter $\w_t$ may be infinite-dimensional. This added expressiveness brings a fundamental algorithmic difficulty: The standard importance-weighted loss estimator from linear bandits can take values of unbounded magnitude when the feature map lies in an infinite-dimensional RKHS, which rules out a direct application of linear-bandit techniques.

To make this concrete, we briefly recall the Exp2 algorithm developed for linear bandits. At each round $t$, Exp2 maintains a distribution $p_t$ over the action set, samples $\x_t \sim p_t$, and observes the scalar loss $y_t = \inner{\w_t}{\x_t}$. From this single observation, it constructs the importance-weighted estimator
\[
    \widehat{\w}_t \;=\; y_t\, \Sigma_t^{-1}\x_t,
    \qquad
    \Sigma_t \;=\; \E_{\x \sim p_t}[\x \x^\top].
\]
Exp2 then updates $p_t$ by exponential weights on the loss proxy $\hat{\ell}_t(\x) := \inner{\widehat{\w}_t}{\x}$. The Exp2 algorithm requires a positive lower bound on the minimum eigenvalue of $\Sigma_t$, which is ensured by mixing $p_t$ with a suitable exploration distribution (e.g., John's distribution), and yields the optimal regret $\widetilde{\O}(\sqrt{d\, T \log|\X|})$~\citep{conf/nips/DaniHK07,bubeck2012towards}. Translating this construction to the kernel setting amounts to replacing $\x$ by $\Phi(\x)$ throughout, with $\Sigma_t = \E_{\x \sim p_t}[\Phi(\x)\Phi(\x)^\top]$ now acting on the RKHS $\H$. However, in an infinite-dimensional RKHS, the covariance operator $\Sigma_t$ may have eigenvalues arbitrarily close to zero, which can make the loss estimator unbounded.

\paragraph{Existing biased estimators and limitations.}
To circumvent this issue, existing work~\citep{chatterji2019online,neu2024go} resorts to biased estimators that trade some bias for variance control. \citet{chatterji2019online} construct a finite-dimensional proxy kernel by projecting the Mercer expansion of $\ke$ to its top-$m$ eigenfunctions, and run Exp2 in this $m$-dimensional space. While the truncation restores invertibility of the covariance and makes Exp2 applicable, the bias it introduces is hard to balance tightly against the estimator variance, leaving room for improvement in the resulting bound. When the kernel satisfies a polynomial eigenvalue decay condition with exponent $\beta>1$, their analysis yields a regret bound of order $\widetilde{\O}(T^{\beta/(2(\beta-1))})$. This gives a suboptimal rate for Matérn kernels and leaves a gap to the lower bound $\Omega(T^{(\beta+1)/(2\beta)})$ established in the same paper. Moreover, to control the bias introduced by their projection-based estimator,~\citet{chatterji2019online} restrict the adversary to rank-one losses $\w_t=\Phi(\z_t)$, so that the loss can be written as $\ell_t(\x)=\ke(\x,\z_t)$. Our approach removes this restriction and handles arbitrary bounded RKHS losses $\w_t\in\mathcal \H$ with $\Vert\w_t\Vert_{\mathcal H}\le B$. 

As outlined in the introduction, \citet{neu2024go} consider the kernelized contextual bandit setting and design a different biased estimator via Kernel Geometric Resampling, a finite-step truncation of a Neumann-series approximation to $\Sigma_t^{-1}$. Even with this approximation, the small eigenvalues of $\Sigma_t$ still leave the loss estimator with potentially large magnitude, so they replace exponential weights with a FTRL update under log-barrier regularization, which is more robust to unbounded loss estimators. Their bound matches the optimal $T$-exponent $T^{(\beta+1)/(2\beta)}$ under polynomial eigenvalue decay. Although their setting is not the same as ours, their estimator can also be applied to our setting, but the use of log-barrier inherently incurs at least a $\sqrt{|\X|}$ dependence on the action-set cardinality, which is highly unfavorable in the kernel bandit setting as we discussed at the end of Section \ref{sec:related}.

\subsection{Our Algorithm}

Our method follows the Exp2 template with two key modifications targeting the obstacles identified above: a \emph{regularized estimator} that keeps the loss estimator bounded despite the vanishing eigenvalues of $\Sigma_t$, and a \emph{correction term} that compensates for the bias introduced by regularization.

\paragraph{Regularized estimator.}
We replace the design covariance $\Sigma_t = \E_{\x \sim p_t}[\Phi(\x)\Phi(\x)^\top]$ with its ridge-regularized version $\Sigma_t^\lambda := \Sigma_t + \lambda I_{\H}$ and form the importance-weighted estimator
\begin{equation}\label{eq:reg-estimator}
    \widehat{\w}_t \;=\; y_t \cdot (\Sigma_t^\lambda)^{-1}\Phi(\x_t),
    \qquad
    y_t = \inner{\Phi(\x_t)}{\w_t}_{\H}.
\end{equation}
The added $\lambda I_{\H}$ term enforces a positive lower bound on the spectrum of $\Sigma_t^\lambda$, keeping $\widehat{\w}_t$ bounded even when $\Sigma_t$ has eigenvalues close to zero; the price paid for this is bias. Letting $\mathcal{F}_{t-1}$ denote the filtration of the history before round $t$, a direct calculation (deferred to \pref{lem:exp2-cond-mean} in the appendix) shows that the conditional bias of $\widehat{\w}_t$ takes the explicit form
\begin{equation}\label{eq:bias-form}
    \w_t - \E[\widehat{\w}_t \mid \mathcal{F}_{t-1}] \;=\; \lambda\,(\Sigma_t^\lambda)^{-1}\w_t.
\end{equation}
This explicit form is what the correction term introduced next will be designed to cancel.

\begin{algorithm}[!t]
\caption{Exponential Weights with Regularized Estimator}
\label{alg:exp2}
\begin{algorithmic}[1]
\REQUIRE Action set $\X$, kernel $\ke$, learning rate $\eta > 0$,
         mixing coefficient $\gamma \in (0,1]$,
         regularization coefficient $\lambda > 0$.

\STATE \textbf{Initialize:} Let $q_1$ be a uniform distribution over $\X$, and define the exploration distribution
\[
    \nu_G \in \argmin_{\nu\in\Delta(\X)}
    \max_{\x\in\X}
    \norm{\Phi(\x)}^2_{(\Sigma(\nu)+(\lambda/\gamma) I_{\H})^{-1}}.
\]
\FOR{$t = 1,2,\ldots,T$}
    \STATE Set $p_t = (1-\gamma)q_t + \gamma \nu_G$, draw $\x_t \sim p_t$, and observe $y_t = \inner{\Phi(\x_t)}{\w_t}_{\H}$.
    \STATE Compute the regularized estimator $\widehat{\w}_t$ via \pref{eq:reg-estimator}. 
    \STATE  Construct the corrected loss proxy $\widehat{\ell}_t$ via \pref{eq:loss-proxy}.
    \STATE Update $q_{t+1}(\x) \propto q_t(\x)\exp\!\big(-\eta\,\widehat{\ell}_t(\x)\big)$ for all $\x \in \X$.
\ENDFOR
\end{algorithmic}
\end{algorithm}

\paragraph{Correction term.}
To compensate for this regularization bias, we define the loss proxy
\begin{equation}\label{eq:loss-proxy}
    \widehat{\ell}_t(\x) \;=\; \inner{\widehat{\w}_t}{\Phi(\x)}_{\H} - c_t(\x),
    \qquad
    c_t(\x) \;=\; B\sqrt{\lambda}\, \norm{\Phi(\x)}_{(\Sigma_t^\lambda)^{-1}}.
\end{equation}
The specific form of the correction term $c_t$ comes directly from the bias formula~\eqref{eq:bias-form}. Applying Cauchy--Schwarz together with $\|\w_t\|_\H \le B$ to the right-hand side of~\eqref{eq:bias-form} yields, for every action $\x$,
\[
    \bigl|\inner{\Phi(\x)}{\w_t - \E[\widehat{\w}_t \mid \mathcal{F}_{t-1}]}_{\H}\bigr|
    \;\le\;
    B\sqrt{\lambda}\,\norm{\Phi(\x)}_{(\Sigma_t^\lambda)^{-1}}
    \;=\; c_t(\x),
\]
so $c_t(\x)$ is a pointwise upper bound on the bias of $\inner{\widehat{\w}_t}{\Phi(\x)}_\H$, and subtracting it acts as a debiasing adjustment that ensures $\widehat{\ell}_t(\x)$ underestimates the true loss in expectation. Specifically, as detailed in \pref{lem:exp2-comp-bias}, $c_t$ offsets the comparator-side bias
$B_T^{\x_*} := \E\bigl[\sum_{t=1}^T \inner{\Phi(\x_*)}{\w_t - \widehat{\w}_t}_\H\bigr]$
and makes the resulting residual non-positive, leaving only the learner-side bias, which is more tractable to bound. 

We note that the use of a correction term arises in various online-learning settings \citep{ICML'14:adaptivity-optimism, JMLR'24:Sword++}, where it serves to translate a bias (or error) term from the comparator side to the learner side of the regret decomposition. In the context of adversarial linear/kernel bandits and policy optimization, different forms of correction have been designed to compensate for the bias of different loss estimators~\citep{luo2021policy, liu2023bypassing,neu2024go}. Our correction arises naturally from the closed-form bias of the regularized loss estimator and integrates seamlessly into the exponential-weights update without inflating the order of the loss-estimator range, which is essential for the regret analysis.

\paragraph{Overall algorithm.} Combining the two ingredients above with a mixing of exploration distribution and exponential-weights update yields \pref{alg:exp2}. At each round $t$, the learner samples $\x_t \sim p_t = (1-\gamma)q_t + \gamma \nu_G$, a mixture of the current distribution $q_t$ and a fixed exploration distribution $\nu_G$ given by a regularized G-optimal design (e.g., see~\citep{camilleri21a}). Upon observing $y_t$, it forms the corrected loss proxy $\widehat{\ell}_t$ via~\eqref{eq:reg-estimator} and~\eqref{eq:loss-proxy} and updates $q_{t+1}$ by exponential weights on $\widehat{\ell}_t$. Although defined through $\widehat{\w}_t \in \H$, the proxy $\widehat{\ell}_t(\x)$ admits a closed-form expression in terms of kernel evaluations on $\X$, so $\widehat{\w}_t$ is never instantiated explicitly. We defer the computational details to the end of this section. 

\begin{myRemark}[Computational Cost]
The computational cost of \pref{alg:exp2} mainly comes from two parts: evaluating the loss proxy $\widehat{\ell}_t$ at each round, and computing the exploration distribution $\nu_G$ once at initialization. Both can be implemented entirely via kernel evaluations $\ke(\x,\y) = \inner{\Phi(\x)}{\Phi(\y)}_\H$, without ever instantiating the feature map $\Phi(\cdot)$ explicitly. The two terms in the loss proxy $\widehat{\ell}_t(\x) = \inner{\widehat{\w}_t}{\Phi(\x)}_\H - c_t(\x)$ are both bilinear forms $\Phi(\x)^\top(\Sigma_t^\lambda)^{-1}\Phi(\y)$ in the regularized covariance. By \citet[Lemma 1]{camilleri21a}, any such form admits the closed-form expression
$
    \Phi(\x)^\top(\Sigma_t^\lambda)^{-1}\Phi(\y)
    \;=\;
    \tfrac{1}{\lambda}\bigl(k(\x,\y) - k_t(\x)^\top(K_t + \lambda I_{|\X|})^{-1}\, k_t(\y)\bigr),
$
where $[K_t]_{i,j} = \sqrt{p_t(\x_i)\,p_t(\x_j)}\,k(\x_i,\x_j)$ and $[k_t(\x)]_i = \sqrt{p_t(\x_i)}\,k(\x_i,\x)$ are both expressed only through kernel evaluations on $\X$. The dominant per-round cost is the inversion of $K_t + \lambda I_{|\X|}$ at $\O(|\X|^3)$. Once this is computed, each evaluation of $\widehat{\ell}_t(\x)$ takes $\O(|\X|^2)$. This complexity might ideally be brought down to at most linear in $|\X|$ (the cost of iterating over the arms), but we leave this for future work. In practice, since our estimator follows the same structure as kernel ridge regression (KRR), acceleration techniques, such as Nystr\"{o}m~\citep{journals/jmlr/DrineasM05} or random Fourier feature~\citep{conf/icml/AvronKMMVZ17} approximations, can also be applied.

The exploration distribution $\nu_G \in \arg\min_{\nu\in\Delta(\X)}\max_{\x\in\X}\norm{\Phi(\x)}^2_{(\Sigma(\nu)+(\lambda/\gamma)I_\H)^{-1}}$ depends only on the fixed parameters $\lambda$ and $\gamma$, so it is computed once at initialization. The objective is convex in $\nu$ and its gradients are again bilinear forms of the same kind, so $\nu_G$ can be obtained by first-order methods using only kernel evaluations~\citep{camilleri21a}.
\remarkend
\end{myRemark}

\subsection{Regret Guarantees}
The following theorem states the regret guarantee of \pref{alg:exp2}, and will be proved below.

\begin{restatable}{myThm}{maintheorem}
\label{thm: main}
    Under \pref{assum:bounded-kernel}, let $\eta>0$ and $\gamma,\lambda \in(0,1]$. Define the effective dimension as $d_*(\rho) = \max_{\nu\in\Delta_{\X}} \mathrm{Tr}\left((\Sigma(\nu)+\rho I_{\H})^{-1}\Sigma(\nu)\right)$, where $\Sigma(\nu) = \E_{\x\sim \nu}[\Phi(\x)\Phi(\x)^\top]$. If $\lambda\leq 1/(2\eta B)$ and $d_*(\lambda/\gamma)\leq \gamma / (2\eta B)$, Algorithm~\ref{alg:exp2} with uniform initialization $q_1$ guarantees
    \begin{align*}
    \Reg_T
    \leq
    \frac{\log \vert \X\vert}{\eta}
    + 2(1+\lambda)B^2\eta d_*\left( \lambda\right) T      + 4B\sqrt{
        \lambda T^2 d_*(\lambda)
    }
    + B\sqrt{
        \lambda\gamma T^2\,
        d_{*}(\lambda/\gamma)
    }+ 2 \gamma BT 
,
\end{align*}
\end{restatable}

With an appropriate choice of parameters, our algorithm enjoys the following regret guarantee in terms of the effective dimension. The proof of this corollary and the following kernel-specific rates are deferred to \pref{app:cor1}.

\begin{restatable}{myCor}{maincor}
\label{cor: maincor}
Under \pref{assum:bounded-kernel} , with the choices ,
% $
%     \bar{\gamma}_T(\alpha)
%     :=
%     \max_{\x_1,\ldots,\x_T\in\mathcal X}
%     \log\det\left(
%         I+\frac{1}{\alpha}
%         \sum_{t=1}^T
%         \Phi(x_t)\Phi(x_t)^\top
%     \right).
% $
% Let
\[
    \lambda=\frac{1}{T},
    \qquad
    \eta
    =
    \frac{1}{2B}
    \sqrt{
        \frac{\log(e|\mathcal X|)}
        {2(1+\lambda)d_*(\lambda)T}
    },
    \qquad
    \gamma
    =
    \min \left\{  
    \sqrt{
        \frac{2d_*(\lambda) \log(e|\mathcal X|)}{(1+\lambda)T}
    },1\right\},
\]
\pref{alg:exp2} satisfies
$
    \Reg_T
    =
    \mathcal{O}
    \left(
        B\sqrt{d_*(\lambda) T\log(|\mathcal X|)}
    \right).
$
\end{restatable}

In general, the effective dimension $d_*$ may be difficult to evaluate or bound directly; we proceed to detail two approaches to doing so and attaining more specific regret bounds.

\paragraph{Bounding $d_*$ via maximum information gain.}
The effective dimension $d_*(\lambda)$ used in \pref{cor: maincor} can be
related to the maximum information gain widely studied in the stochastic
kernel bandit literature. Specifically, define
\[
    \bar{\gamma}_T
    \;:=\; 
    \max_{\x_1,\ldots,\x_T \in \X}
    \frac{1}{2}\log\det\!\left(I_\H + \sum_{t=1}^T \Phi(\x_t)\Phi(\x_t)^\top\right),
\]
which corresponds to the maximum information gain of \citet{srinivas2010gp}
at noise level $\sigma^2 = 1$. \pref{lem:effdim-vs-infogain} in
\pref{subsec:exp2-lemmas} shows that $d_*(1/T) \le \O(\bar{\gamma}_T)$, so
the regret bound of \pref{cor: maincor} translates to
\[
    \Reg_T = \widetilde{\O}\!\left(B\sqrt{\bar{\gamma}_T\, T\, \log|\X|}\right).
\]
In particular, when $B$ is constant and $\log|\X| = O(\log T)$,\footnote{This condition holds when the dimension is fixed and we let $|\X| = T^{d/2}$ in the covering argument described at the start of Section~\ref{sec:problem_setup}.} this simplifies to $\widetilde{\O}(\sqrt{T\bar{\gamma}_T})$, matching the $\widetilde{\O}(\sqrt{T\bar{\gamma}_T})$ rate achieved by several stochastic kernel bandit algorithms~\citep{valko2013finite,salgia2021domain,li2022gaussian}. Moreover, bounds on the maximum information gain are known for commonly used kernels on a compact subset of $\mathbb{R}^d$ (e.g., $[0,1]^d$) such as Mat\'ern and SE~\citep{vakili2021information}, which lead to the following kernel-specific rates:
\begin{itemize}[leftmargin=5ex]
    \item \textbf{Mat\'ern$(\nu,d)$ kernel:} we have $\bar{\gamma}_T = \widetilde{\O}(T^{d/(2\nu+d)})$, giving $\Reg_T = \widetilde{\O}(T^{(\nu+d)/(2\nu+d)})$.
    \item \textbf{SE kernel:} we have $\bar{\gamma}_T = \O((\log T)^{d+1})$, giving $\Reg_T = \widetilde{\O}(\sqrt T)$.
\end{itemize}
The concurrent work of \citet{iwazaki2026nearly} establishes lower bounds for both of these kernels, showing that our upper bound is tight up to logarithmic factors in $T$. Both rates also match the optimal regret rates known for stochastic kernel bandits; see, e.g., \citet{vakili2021information,li2022gaussian}. 

\begin{myRemark} While the adversarial kernel bandit problem intuitively seems to be more difficult than the stochastic counterpart, we highlight that neither is a special case of the other.  In particular, the additive noise in the stochastic problem can create bottlenecks that are not present in the adversarial problem, and for linear rewards these can actually lead to strictly higher regret; see for example \cite[Chapter 29]{lattimore2020bandit}.
\remarkend
\end{myRemark}

\paragraph{Bounding $d_*$ via eigenvalue decay conditions.} While relating the effective dimension to the maximum information gain (which has been studied extensively previously) is a natural approach to obtaining results for specific kernels, another route is to place assumptions directly on the kernel eigenvalue decay and upper bound $d_*(\lambda)$ directly based on that decay, with suitable tuning of
$\eta$. Corollaries 1 and 2 below carry this out for polynomial and exponential eigenvalue decay, respectively, defined as follows.

\begin{myDef}[Eigenvalue Decay]
\label{def:decay_1}
Let $p$ be any probability measure over $\X$, and let
$\mu_{1} \ge \mu_{2} \ge \cdots \ge 0$ denote the eigenvalues of the covariance operator
$
    \Sigma
    =
    \E_{\x\sim p}\!\left[\Phi(\x) \Phi(\x)^\top\right].
$
\begin{enumerate}[leftmargin=4ex]
    \item $\ke$ is said to have $(C,\beta)$-polynomial eigenvalue decay, with $\beta>1$, if there exists a constant $C>0$ such that
    $
        \mu_j \leq C j^{-\beta}
    $
    for all integers $j\ge 1$, regardless of $p$.

    \item $\ke$ is said to have $(C,\beta)$-exponential eigenvalue decay if there exist constants $C>0$ and $\beta>0$ such that
    $
        \mu_j \leq C e^{-\beta j}
    $
    for all integers $j\ge 1$, regardless of $p$.
\end{enumerate}
\end{myDef}

It is worth noting that the uniformity over $p$ makes this definition more restrictive than the assumption of similar decay for a fixed measure (e.g., see \cite[Definition 11]{chatterji2019online}).  Nevertheless, under these conditions, we have the following corollaries, whose proofs are deferred to \pref{app:decay}.

\begin{restatable}{myCor}{maincorollary}
\label{cor: main}
Suppose $\ke$ has $(C, \beta)$-polynomial eigenvalue decay defined in \pref{def:decay_1}. Under \pref{assum:bounded-kernel} , Algorithm~\ref{alg:exp2} initialized with the uniform distribution
$q_1$ over $\X$ and run with
\[
m = \left\lceil T^{1/\beta}\right\rceil, 
\quad
    \lambda = m^{-\beta},
    \quad
    \gamma = m^{-\frac{\beta-1}{2}},
    \quad
    \eta = \frac{1}{2B(1+C/(\beta-1))}\,m^{-\frac{\beta+1}{2}}
\]
guarantees
\[
\Reg_T
=
\mathcal{O}\!\left(
B\left(
1+\left(1+\frac{C}{\beta-1}\right)\log|\X|+\sqrt{1+\frac{C}{\beta-1}}
\right)
T^{\frac{\beta+1}{2\beta}}
\right).
\]
Thus, when $(B,\beta,C)$ are constant and $\log |\mathcal{X}| = O(\log T)$, we have $\Reg_T = O\big( T^{\frac{\beta+1}{2\beta}} \log T \big)$.
\end{restatable}

\begin{restatable}{myCor}{maincorollarytwo}
\label{cor: main_2}
If $\ke$ has $(C,\beta)$-exponential eigenvalue decay defined in
\pref{def:decay_1}. Under \pref{assum:bounded-kernel}, Algorithm~\ref{alg:exp2}
initialized with the uniform distribution $q_1$ over $\X$ and run with
\[
    m = \left\lceil \frac{\log T}{\beta} \right\rceil,
    \quad
    \lambda = e^{-\beta m}, \quad
    \gamma
    =
    \sqrt{
        \frac{
            \left(m+\frac{C}{e^\beta-1}\right)\log(e|\X|)
        }{T}
    },
    \quad
    \eta
    =
    \frac{\gamma}{
        2B\left(m+\frac{C}{e^\beta-1}\right)
    },
\]
provided that $T$ is large enough so that $\gamma\le 1$, guarantees
\[
    \Reg_T
    =
    \mathcal{O}\!\left(
        B
        \sqrt{
            T
            \left(
                \frac{\log T}{\beta}
                +
                \frac{C}{e^\beta-1}
            \right)
            \log(e|\X|)
        }
    \right).
\]
Thus, when $(B,\beta,C)$ are constant and
$\log|\X| = O(\log T)$, we have
$
    \Reg_T = O\!\big(\sqrt{T\log T}\big).
$
\end{restatable}

\begin{myRemark}
    For the exponential decay case, the regret bound is $\widetilde{\O}(\sqrt{T})$, which matches the tight rate in~\citet{chatterji2019online}. For the polynomial decay case, our regret has the key dependence $\widetilde{O}\big(T^{\frac{\beta+1}{2\beta}}\big)$, which improves over the previous $\widetilde{O}\big(T^{\frac{\beta}{2(\beta-1)}}\big)$ upper bound from~\citet{chatterji2019online}, albeit under the stronger version of the decay assumption as noted above. Although the same paper also proves a lower bound of order  $\Omega\big(T^{\frac{\beta+1}{2\beta}}\big)$, this lower bound does not directly establish the tightness of our result. One obstacle is that their construction uses a $T$-dependent action set, making the $\log|\mathcal X|$ factor in our upper bound no longer lower order.  Moreover, the result relies on a different eigendecay condition and adversary model, making it incompatible with our setting. We provide a more detailed discussion of these issues in Appendix~\ref{app:discussion_1}.
\end{myRemark}

%------------------------------------------------------------------
\section{Regret Analysis}

\label{sec: regret_analysis}
In this section, we provide the regret analysis that establishes \pref{thm: main}.  The proofs of some auxiliary lemmas are deferred to the appendices.

% \maintheorem*
\subsection{Proof of Theorem~\ref{thm: main}}
\begin{proof}[Proof of Theorem~\ref{thm: main}]
\label{subsec:exp2-analysis}
%------------------------------------------------------------------
Let $p_* \in \Delta(\X)$ denote the probability distribution that puts all its mass on $\x_*$, i.e., $p_*(\x)=\mathbf{1}\{\x=\x_*\}$, where $\x_* \in \arg\min_{\x\in\X} \sum_{t=1}^T \inner{\Phi(\x)}{\w_t}_{\H}$ is a best fixed action in hindsight. We also define $\F_{t-1} = \sigma(\x_1,\w_1,\ldots,\x_{t-1},\w_{t-1})$ as the history available before round $t$. Since both $\w_t$ and the learner's prediction distribution $p_t$ are $\F_{t-1}$-measurable, and $\x_t$ is sampled conditionally on $\F_{t-1}$ according to $p_t$, i.e., $\mathbb P(\x_t=\x\mid \F_{t-1})=p_t(\x),\forall \x\in\X,$ applying the tower property of conditional expectation yields
\begin{align}
\Reg_T
&:=
\E\!\left[
\sum_{t=1}^T \inner{\Phi(\x_t)}{\w_t}_{\H}
-
\sum_{t=1}^T \inner{\Phi(\x_*)}{\w_t}_{\H}
\right] \notag\\
&=
\E\!\left[
\sum_{t=1}^T
\E\!\left[
\inner{\Phi(\x_t)}{\w_t}_{\H}
\,\middle|\, \F_{t-1}
\right]
-
\sum_{t=1}^T \inner{\Phi(\x_*)}{\w_t}_{\H}
\right] \notag\\
&=
\E\!\left[
\sum_{t=1}^T
\sum_{\x\in\X}
p_t(\x)\inner{\Phi(\x)}{\w_t}_{\H}
-
\sum_{t=1}^T
\sum_{\x\in\X}
p_*(\x)\inner{\Phi(\x)}{\w_t}_{\H}
\right] \notag\\
&=
\E\!\left[
\sum_{t=1}^T
\sum_{\x\in\X}
\bigl(p_t(\x)-p_*(\x)\bigr)
\inner{\Phi(\x)}{\w_t}_{\H}
\right].
\label{eq:exp2-reg-tower}
\end{align}

We next relate the loss function induced by $\w_t$ to the loss estimator used in Algorithm~\ref{alg:exp2}. Recall the choice $c_t(\x) := B\sqrt{\lambda} \Vert \Phi(\x)\Vert_{(\SigLam)^{-1}}$, so that
 $
    \hat{\ell}_t(\x) = \inner{\Phi(\x)}{\wht}_\H - c_t(\x).$
Then
\[
    \inner{\Phi(\x)}{\w_t}_\H
    \;=\;
    \hat{\ell}_t(\x)
    + \inner{\Phi(\x)}{\w_t-\wht}_\H
    + c_t(\x),
\]
and separating contributions yields the following \emph{decomposition with correction
terms}:
\begin{align}
    \Reg_T
    \;=\;
    &\underbrace{
        \E\!\left[\sum_{t=1}^T \sum_{\x\in\X}\bigl(p_t(\x)-p_*(\x)\bigr)
            \hat{\ell}_t(\x)\right]
    }_{\displaystyle\Reg_T^{\hat{\ell}}}
    \notag\\[4pt]
    &\quad+\;
    \underbrace{
        \E\!\left[\sum_{t=1}^T \sum_{\x\in\X}p_t(\x)\inner{\Phi(\x)}{\w_t-\wht}_\H\right]
    }_{\displaystyle B_T^{\x_t}}+\;
    \underbrace{
        \E\!\left[\sum_{t=1}^T \sum_{\x\in\X}p_t(\x)\,c_t(\x)\right]
    }_{\displaystyle C_T^{\x_t}}
    \notag\\[4pt]
    &\quad
    \;-\;
    \underbrace{
        \E\!\left[\sum_{t=1}^T \inner{\Phi(\x_*)}{\w_t-\wht}_\H\right]
    }_{\displaystyle B_T^{\x_*}}
    \;-\;
    \underbrace{
        \E\!\left[\sum_{t=1}^T c_t(\x_*)\right]
    }_{\displaystyle C_T^{\x_*}}.
    \label{eq:exp2-reg-decomp}
\end{align}

In the above, $\Reg_T^{\hat{\ell}}$ is the regret term defined in terms of the loss estimator. The terms $B_T^{\x_t}$ and $B_T^{\x_*}$ are the regularization-induced bias terms on the learner side and comparator side, respectively. The terms
$C_T^{\x_t}$ and $C_T^{\x_*}$ are correction terms chosen to match these
biases. This structural choice for the correction term is crucial: it ensures that the comparator-side
residual $-(B_T^{\x_*}+C_T^{\x_*})$ is non-positive, rather than forcing us to
use a looser absolute-value bound for $B_T^{\x_*}$. The learner-side bias, comparator-side bias, and estimated-loss regret are then
controlled by \pref{lem:exp2-learn-bias}, \pref{lem:exp2-comp-bias}, and
\pref{lem:exp2-reg-ew}, respectively. The detailed proofs of these lemmas are
deferred to \pref{app:proof-exp2}.

\begin{restatable}[Learner-side Residual]{myLemma}{learnerresidual}
\label{lem:exp2-learn-bias}
Under~\pref{assum:bounded-kernel}, for any $\lambda>0$, the learner-side residual satisfies
\begin{align}
    B_T^{\x_t}+C_T^{\x_t}
    \;\le\;
    4B\sqrt{
        \lambda T\,
        \E\!\left[
            \sum_{t=1}^T d_{\mathrm{eff}}(\Sigma_t,\lambda)
        \right]
    } .
    \label{eq:exp2-learn-bias-bound}
\end{align}
Here, for each $t\in[T]$, $d_{\mathrm{eff}}(\Sigma_t,\lambda)
    :=
    \Tr\!\bigl((\Sigma_t^\lambda)^{-1}\Sigma_t\bigr)$
    denotes the $p_t$-dependent version of effective dimension, where
$    \Sigma_t
    =
    \E_{\x\sim p_t}[\Phi(\x)\Phi(\x)^\top]$ and 
    $\Sigma_t^\lambda
    =
    \Sigma_t+\lambda I_{\H}.$
\end{restatable}
%------------------------------------------------------------------

\begin{restatable}[Comparator-side Residual]{myLemma}{comparatorresidual}
\label{lem:exp2-comp-bias}
Under \pref{assum:bounded-kernel} and for any $\lambda > 0$, the comparator-side residual is
non-positive: 
\begin{equation}
    - B_T^{\x_*} - C_T^{\x_*}
    \;\le\;
    0.
    \label{eq:exp2-comp-bias-bound}
\end{equation}
\end{restatable}

\begin{restatable}[Regret Bound via EW]{myLemma}{estimatedregret}
\label{lem:exp2-reg-ew}
Under \pref{assum:bounded-kernel}, for any $\lambda > 0$, $\eta > 0$,
and $\gamma \in (0,1]$, let
\[
    \nu_D
    \;\in\;
    \argmax_{\nu\in\Delta(\X)}
    \log\det\!\Big(
        \E_{\x\sim\nu}[\Phi(\x)\Phi(\x)^\top] + \tfrac{\lambda}{\gamma} I_\H
    \Big)
\]
be the regularized D-optimal exploration distribution, and define
$\Sigma(\nu_D) := \E_{\x\sim\nu_D}[\Phi(\x)\Phi(\x)^\top]$.  
Under the condition
\begin{equation}
d_{\mathrm{eff}}(\Sigma(\nu_D),\lambda/\gamma)
    \;\le\;
    \frac{\gamma}{2\eta B},
    \label{eq:exp2-ew-condition}
\end{equation}
together with $\lambda \le 1/(2\eta B)$, \pref{alg:exp2} satisfies
\begin{align}
\label{eq:exp2-reg-ew-bound-effective}
\Reg_T^{\hat{\ell}}
\;\le\;&
2\gamma BT + \frac{\log|\X|}{\eta}+ 2(1+\lambda)B^2\eta\cdot\E\left[ \sum_{t=1}^T d_{\mathrm{eff}}(\Sigma_t,\lambda)\right] +  B\sqrt{{\lambda \gamma T}\sum_{t=1}^T d_{\mathrm{eff}}(\Sigma(\nu_D),\lambda/\gamma)}. 
\end{align}
\end{restatable}
Combining \eqref{eq:exp2-reg-decomp} with \pref{lem:exp2-learn-bias},
\pref{lem:exp2-comp-bias} and~\pref{lem:exp2-reg-ew}, we obtain
\begin{align*}
    \Reg_T \leq{}& 2 \gamma BT + \frac{\log \vert \X\vert}{\eta} + 2(1+\lambda)B^2\eta\cdot\E\left[ \sum_{t=1}^T d_{\mathrm{eff}}(\Sigma_t,\lambda)\right]\\
    {}& +B\sqrt{{\lambda \gamma T}\sum_{t=1}^T d_{\mathrm{eff}}(\Sigma(\nu_D),\lambda/\gamma)}  +   4B\sqrt{
        \lambda T\,
        \E\!\left[
            \sum_{t=1}^T d_{\mathrm{eff}}(\Sigma_t,\lambda)
        \right]
    }.
\end{align*}
Defining $d_*(\rho) = \max_{\nu\in\Delta_{\X}} \mathrm{Tr}\left((\Sigma(\nu)+\rho I_{\H})^{-1}\Sigma(\nu)\right)$, where $\Sigma(\nu) = \E_{\x\sim \nu}[\Phi(\x)\Phi(\x)^\top]$, we then obtain 
    \begin{align*}
    \Reg_T
    \leq{}&
   \frac{\log \vert \X\vert}{\eta}
    + 2(1+\lambda)B^2\eta d_*\left( \lambda\right) T 
    + B\sqrt{
        \lambda\gamma T^2\,
        d_{*}(\lambda/\gamma)
    }
    + 4B\sqrt{
        \lambda T^2 d_*(\lambda)
    }+ 2 \gamma BT,
\end{align*}
which completes the proof.
\end{proof}

\section{Conclusion}
\label{sec:conclusion}

For the adversarial kernel bandit problem, we proposed an exponential-weights algorithm with a regularized importance-weighted loss estimator and a correction term that mitigates the regularization bias. We established a regret bound of $\widetilde{\O}\big(B\sqrt{T\, d_*(\lambda)\,\log|\X|}\big)$, which matches the optimal rate of the analogous stochastic kernel bandit problem up to logarithmic factors. As a direct consequence, our result yields a near-optimal $\widetilde{\O}\big(T^{(\nu+d)/(2\nu+d)}\big)$ regret bound for the Mat\'ern$(\nu,d)$ kernel, improving over the best-known prior rate of \citet{chatterji2019online} while removing their rank-one adversary assumption. We also derived corollaries under structural assumptions on the kernel spectrum, obtaining $\widetilde{\O}\big(T^{(\beta+1)/(2\beta)}\big)$ regret under polynomial eigenvalue decay with exponent $\beta>1$ and $\widetilde{\O}(\sqrt{T})$ regret under exponential eigenvalue decay.  Possible future directions include obtaining high-probability regret bounds and considering the contextual kernel bandit setting.

\section*{Acknowledgment}

We thank Shogo Iwazaki for helpful comments on an earlier version of our paper, in particular leading to the addition of \pref{lem:effdim-vs-infogain}.  
This work is supported by the Singapore National Research Foundation under its AI Visiting Professorship programme.

% \newpage
\bibliography{myBib}
\bibliographystyle{plainnat}
\newpage
\appendix
\section{Omitted Details from  \pref{sec: regret_analysis}}
\label{app:proof-exp2}

%------------------------------------------------------------------
\subsection{Useful Lemmas}
\label{subsec:exp2-lemmas}
The following lemma characterizes the conditional expectation of the regularized
importance-weighted estimator. It shows that the estimator is biased due to
regularization: instead of recovering $\w_t$ in expectation, it returns a
shrunk version $(\Sigma_t^\lambda)^{-1}\Sigma_t \w_t = (I_{\mathcal{H}} - \lambda(\Sigma_t^\lambda)^{-1}) w_t$.
This explicit form of the bias will be useful in controlling the effect of
regularization in the regret analysis.

\begin{restatable}[Conditional Mean of Loss Estimator]{myLemma}{conditionalmean}
\label{lem:exp2-cond-mean}
Fix any round $t \in [T]$, and let
$
\Sigma_t := \E_{\x\sim p_t}[\Phi(\x)\Phi(\x)^\top]
$ be the unregularized covariance operator. Then for any $\lambda > 0$, $
\Sigma_t^\lambda = \Sigma_t + \lambda I_{\mathcal H}.
$
Conditioning on the history $\mathcal{F}_{t-1}$, the estimator
\[
\wht
=
\inner{\Phi(\x_t)}{\w_t}_{\mathcal H}\,(\Sigma_t^\lambda)^{-1}\Phi(\x_t), \quad (\x_t\, \vert\, \F_{t-1}) \sim p_t,
\]
satisfies
\begin{equation}
\E[\wht \mid \mathcal{F}_{t-1}]
=
(\Sigma_t^\lambda)^{-1}\Sigma_t\,\w_t
=
\bigl(I_{\mathcal H} - \lambda(\Sigma_t^\lambda)^{-1}\bigr) \w_t.
\label{eq:exp2-cond-exp-what}
\end{equation}
\end{restatable}

\begin{proof}
We condition on $\mathcal{F}_{t-1}$, meaning that $p_t$, $\w_t$, $\Sigma_t$, and
$(\Sigma_t^\lambda)^{-1}$ are fixed. Thus,
\[
\E[\wht \mid \mathcal{F}_{t-1}]
=
(\Sigma_t^\lambda)^{-1}
\E_{\x\sim p_t}\!\left[
    \inner{\Phi(\x)}{\w_t}_{\mathcal H}\,\Phi(\x)
\right].
\]
By the definition of $\Sigma_t$,
\[
\E_{\x\sim p_t}\!\left[
    \inner{\Phi(\x)}{\w_t}_{\mathcal H}\,\Phi(\x)
\right]
=
\E_{\x\sim p_t}\!\left[
    \Phi(\x)\Phi(\x)^\top
\right]\w_t
=
\Sigma_t \w_t,
\]
which implies
\[
\E[\wht \mid \mathcal{F}_{t-1}]
=
(\Sigma_t^\lambda)^{-1}\Sigma_t \w_t.
\]

Finally, since $\Sigma_t^\lambda = \Sigma_t + \lambda I_{\mathcal H}$, we have
\[
(\Sigma_t^\lambda)^{-1}\Sigma_t
=
(\Sigma_t^\lambda)^{-1}(\Sigma_t^\lambda - \lambda I_{\mathcal H})
=
I_{\mathcal H} - \lambda(\Sigma_t^\lambda)^{-1}.
\]
Substituting this identity gives $\E[\wht \mid \mathcal{F}_{t-1}]
=
\bigl(I_{\mathcal H} - \lambda(\Sigma_t^\lambda)^{-1}\bigr)\w_t$ as claimed.
\end{proof}
%------------------------------------------------------------------

The following lemma establishes a uniform bound on the corrected proxy loss. This boundedness condition allows us to invoke the standard exponential-weights regret guarantee despite the estimator being defined in a potentially infinite-dimensional RKHS.
\begin{restatable}[Bounded Loss Estimator]{myLemma}{boundedlossestimator}
\label{lem:exp2-bounded}
Under Assumption~\ref{assum:bounded-kernel} and condition~\eqref{eq:exp2-ew-condition}, and assuming $\lambda \le 1/(2\eta B)$, the loss proxy $\widehat{\ell}_t(\x)$
satisfies $\vert\eta \widehat{\ell}_t(\x)| \le 1.$
\end{restatable}
\begin{proof}
By the definition of the loss proxy, for any $\x \in \X$ we have
\begin{align}
|\hat{\ell}_t(\x)| 
&\le |\inner{\widehat{\w}_t}{\Phi(\x)}_{\H}| + B\sqrt{\lambda}\, \norm{\Phi(\x)}_{(\Sigma_t^\lambda)^{-1}} \notag\\
&\le \norm{{\w}_t}_{\H} \,\norm{\Phi(\x)}_{(\Sigma_t^\lambda)^{-1}} \norm{\Phi(\x_t)}_{(\Sigma_t^\lambda)^{-1}} 
      + B\sqrt{\lambda}\, \norm{\Phi(\x)}_{(\Sigma_t^\lambda)^{-1}} \notag\\
&\le B \, \norm{\Phi(\x)}_{(\Sigma_t^\lambda)^{-1}} \, \norm{\Phi(\x_t)}_{(\Sigma_t^\lambda)^{-1}} 
      + B\sqrt{\lambda}\, \norm{\Phi(\x)}_{(\Sigma_t^\lambda)^{-1}}, \label{eq:bounded-loss-2}
\end{align}
where the second inequality holds using the definition $\wht = \inner{\Phi(\x_t)}{\w_t} (\Sigma_t^\lambda)^{-1}\Phi(\x_t)$, and applying  $\vert \inner{\wht}{\Phi(\x)}\vert =\vert \inner{\Phi(\x_t)}{\w_t} \cdot \inner{\Phi(\x)}{(\Sigma_t^{\lambda})^{-1}\Phi(\x_t)}\vert \leq \norm{{\w}_t}_{\H} \,\norm{\Phi(\x)}_{(\Sigma_t^\lambda)^{-1}} \norm{\Phi(\x_t)}_{(\Sigma_t^\lambda)^{-1}} $ by Assumption~\ref{assum:bounded-kernel} and two applications of Cauchy-Schwarz. The last inequality uses $\norm{\w_t}_{\H}\leq B$, again by Assumption~\ref{assum:bounded-kernel}.

Next, we derive a uniform upper bound on 
$\norm{\Phi(\x)}_{(\Sigma_t^\lambda)^{-1}}$. Recall that Algorithm~\ref{alg:exp2}
uses the exploration distribution
\[
    \nu_G \in \argmin_{\nu\in\Delta(\X)} M_{\lambda/\gamma}(\nu),
\]
where $M_{\lambda/\gamma}(\nu)
    :=
    \max_{\x\in\X}
    \norm{\Phi(\x)}^2_{\left(\Sigma(\nu)+\frac{\lambda}{\gamma}I_{\H}\right)^{-1}}$ and $\Sigma(\nu)
    :=
    \E_{\z\sim\nu}\!\left[\Phi(\z)\Phi(\z)^\top\right].
$

By \citet[Lemma 3]{camilleri21a}, we can relate $M_{\lambda/\gamma}$ to the effective dimension as follows:
\begin{equation}M_{\lambda/\gamma}(\nu_G)
\le \mathrm{Tr}\Big((\Sigma(\nu_D) + \frac{\lambda}{\gamma} I_\H)^{-1} \Sigma(\nu_D)\Big)
= d_{\mathrm{eff}}(\Sigma(\nu_D), \lambda/\gamma), \label{eq:max-Phi}
\end{equation}
where $\nu_D \in \argmax_{\nu\in\Delta(\X)} \log\det\!\big(\E_{\x\sim\nu}[\Phi(\x)\Phi(\x)^\top] + \frac{\lambda}{\gamma} I_\H\big)$. 
Furthermore, since $p_t = (1-\gamma)q_t + \gamma \nu_G$, we have
\[
\Sigma_t^\lambda = \Sigma_t + \lambda I_\H \succeq \gamma \Sigma(\nu_G) + \lambda I_\H = \gamma \Big(\Sigma(\nu_G) + \frac{\lambda}{\gamma} I_\H\Big).
\]
By monotonicity of matrix inversion, we have
\[
(\Sigma_t^\lambda)^{-1} \preceq \frac{1}{\gamma} \big(\Sigma(\nu_G) + \frac{\lambda}{\gamma} I_\H\big)^{-1},
\]
which implies for all $\x \in \X$ that
\begin{equation}
\norm{\Phi(\x)}^2_{(\Sigma_t^\lambda)^{-1}}
\le \frac{1}{\gamma} M_{\lambda/\gamma}(\nu_G)
\le \frac{1}{\gamma} d_{\mathrm{eff}}(\Sigma(\nu_D), \lambda/\gamma). \label{eq:bounded-loss-1}
\end{equation}

Finally, substituting~\eqref{eq:bounded-loss-1} into~\eqref{eq:bounded-loss-2}, we obtain
\[
|\hat{\ell}_t(\x)| 
\le \frac{B}{\gamma} d_{\mathrm{eff}}(\Sigma(\nu_D), \lambda/\gamma)
+ B \sqrt{\frac{\lambda\, d_{\mathrm{eff}}(\Sigma(\nu_D), \lambda/\gamma)}{\gamma}}
\le \frac{1}{2\eta} + \sqrt{\frac{B\lambda}{2\eta}},
\]
where the last inequality follows from condition~\eqref{eq:exp2-ew-condition}.  
Using $\lambda \le 1/(2\eta B)$, it then follows that $|\hat{\ell}_t(\x)| \le \frac{1}{\eta}$ for all $\x \in \X$, and hence
\[
|\eta \hat{\ell}_t(\x)| \le 1.
\]
This completes the proof.
\end{proof}
%------------------------------------------------------------------

The next lemma translates the eigenvalue-decay condition into explicit upper bounds on the effective dimension under both polynomial and exponential decay regimes. These bounds are then used to specialize the general effective-dimension regret bound to each decay setting.

\begin{restatable}{myLemma}{effectivedim}
\label{lem:effective-dim}
Define the effective dimension
$
    d_*(\xi)
    :=
    \max_{p\in\Delta} d_{\mathrm{eff}}(\Sigma(p),\xi),
$
where
$
d_{\mathrm{eff}}(\Sigma(p), \xi)
    :=
    \Tr\!\bigl((\Sigma^\xi(p))^{-1}\Sigma(p)\bigr),
$
$
    \Sigma(p)
    =
    \E_{\x\sim p}[\Phi(\x)\Phi(\x)^\top],
$
and
$
    \Sigma^\xi(p)
    =
    \Sigma(p)+\xi I_{\H}.
$
Then, under the eigenvalue decay conditions of \pref{def:decay_1}, the following hold.
\begin{enumerate}[leftmargin=4ex]
    \item If $\ke$ has $(C,\beta)$-polynomial eigenvalue decay with $\beta>1$, then for any integer $m\ge 1$,
    \[
        d_*(\xi)
        \le
        m+\frac{\epsilon_{\mathrm{poly}}}{\xi},
    \]
    where
    \[
        \epsilon_{\mathrm{poly}}
        :=
        \frac{C}{\beta-1}m^{1-\beta}.
    \]

    \item If $\ke$ has $(C,\beta)$-exponential eigenvalue decay with
    $\beta>0$, then for any integer $m\ge 1$,
    \[
        d_*(\xi)
        \le
        m+\frac{\epsilon_{\mathrm{exp}}}{\xi},
    \]
    where
    \[
        \epsilon_{\mathrm{exp}}
        :=
        \frac{C e^{-\beta(m+1)}}{1-e^{-\beta}}
        =
        \frac{C e^{-\beta m}}{e^\beta-1}.
    \]
\end{enumerate}
\end{restatable}
\begin{proof}
Fix any $p\in\Delta$ and let $\mu_1\ge \mu_2\ge \cdots \ge 0$ denote the eigenvalues of
$\Sigma(p)$. Since $\Sigma^\xi(p)=\Sigma(p)+\xi I_\H$, the operator
$(\Sigma^\xi(p))^{-1}\Sigma(p)$ has eigenvalues
\[
    \frac{\mu_i}{\mu_i+\xi},
    \qquad i\ge 1.
\]
Therefore, for any integer $m\ge 1$,
\begin{align}
d_{\mathrm{eff}}(\Sigma(p),\xi)
    &=
    \Tr\!\bigl((\Sigma^\xi(p))^{-1}\Sigma(p)\bigr) \notag \\
    &=
    \sum_{i=1}^\infty
    \frac{\mu_i}{\mu_i+\xi} \notag \\
    &=
    \sum_{i=1}^m
    \frac{\mu_i}{\mu_i+\xi}
    +
    \sum_{i=m+1}^\infty
    \frac{\mu_i}{\mu_i+\xi} \notag \\
    &\leq
    m
    +
    \frac{1}{\xi}
    \sum_{i=m+1}^\infty \mu_i, \label{eq:effdim-tail}
\end{align}
where we used
\[
    \frac{\mu_i}{\mu_i+\xi}\le 1,
    \qquad
    \frac{\mu_i}{\mu_i+\xi}\le \frac{\mu_i}{\xi}.
\]

Suppose $\ke$ has $(C,\beta)$-polynomial eigenvalue decay with
$\beta>1$. By \pref{def:decay_1}, the bound $\mu_i\le C i^{-\beta}$ holds regardless of $p$, hence
\[
    \sum_{i=m+1}^{\infty}\mu_i
    \le
    \sum_{i=m+1}^{\infty} C i^{-\beta}
    \le
    \int_m^\infty C x^{-\beta}\,dx
    =
    \frac{C}{\beta-1}m^{1-\beta}
    =:
    \epsilon_{\mathrm{poly}}.
\]
Substituting this $p$-independent tail bound into \eqref{eq:effdim-tail} gives
\[
    d_{\mathrm{eff}}(\Sigma(p),\xi)
    \le
    m+\frac{\epsilon_{\mathrm{poly}}}{\xi}
    =
    m+\frac{C}{(\beta-1)\xi}m^{1-\beta},
\]
for every $p\in\Delta$. Taking the maximum over $p\in\Delta$ on the left-hand side yields
\[
    d_*(\xi) = \max_{p\in\Delta} d_{\mathrm{eff}}(\Sigma(p),\xi)
    \le
    m+\frac{\epsilon_{\mathrm{poly}}}{\xi}.
\]

Suppose instead $\ke$ has $(C,\beta)$-exponential eigenvalue decay with
$\beta>0$. By \pref{def:decay_1}, $\mu_i\le C e^{-\beta i}$ regardless of $p$, hence
\[
    \sum_{i=m+1}^{\infty}\mu_i
    \le
    \sum_{i=m+1}^{\infty} C e^{-\beta i}
    =
    C e^{-\beta(m+1)}
    \sum_{k=0}^{\infty} e^{-\beta k}
    =
    \frac{C e^{-\beta(m+1)}}{1-e^{-\beta}}
    =:
    \epsilon_{\mathrm{exp}},
\]
or equivalently,
\[
    \epsilon_{\mathrm{exp}}
    =
    \frac{C e^{-\beta m}}{e^\beta-1}.
\]
Substituting this $p$-independent tail bound into \eqref{eq:effdim-tail} and taking the maximum over $p\in\Delta$, we obtain
\[
    d_*(\xi)
    \le
    m+\frac{\epsilon_{\mathrm{exp}}}{\xi}
    =
    m+
    \frac{C e^{-\beta(m+1)}}{(1-e^{-\beta})\xi}.
\]
This completes the proof.
\end{proof}

The next lemma bounds the second moment of the regularized kernel features appearing in the proxy-loss variance term. This provides the key variance control needed in the exponential-weights regret analysis.
\begin{restatable}{myLemma}{tracebound}
\label{lem:quadratic-leverage}
Fix any round $t\in[T]$ and condition on $\mathcal F_{t-1}$. Let
\[
    \Sigma_t := \mathbb E_{\x\sim p_t}\!\left[\Phi(\x)\Phi(\x)^\top\right],
    \qquad
    \Sigma_t^\lambda := \Sigma_t+\lambda I_{\mathcal H},
\]
and let $A_t := (\Sigma_t^\lambda)^{-1}$. Then
\[
\mathbb E_{\x_t\sim p_t}
\left[
\sum_{\x\in\X}p_t(\x)
\left\langle
\Phi(\x), A_t\Phi(\x_t)
\right\rangle_{\mathcal H}^2
\;\bigg|\;
\mathcal F_{t-1}
\right]
\le
d_{\mathrm{eff}}(\Sigma_t,\lambda),
\]
where $d_{\mathrm{eff}}(\Sigma_t,\lambda)
:=
\operatorname{Tr}\!\left((\Sigma_t+\lambda I_{\mathcal H})^{-1}\Sigma_t\right).$
\end{restatable}

\begin{proof}
We condition on $\mathcal F_{t-1}$, meaning that $p_t$, $\Sigma_t$, and
$A_t=(\Sigma_t+\lambda I_{\mathcal H})^{-1}$ are fixed, and the only
remaining randomness is over $\x_t\sim p_t$. For any fixed $\z\in\X$, by the definition of $\Sigma_t$, we have
\[
\sum_{\x\in\X}p_t(\x)
\left\langle
\Phi(\x),A_t\Phi(\z)
\right\rangle_{\mathcal H}^2
=
\left\langle
A_t\Phi(\z),\Sigma_t A_t\Phi(\z)
\right\rangle_{\mathcal H}.
\]
Taking expectation over $\z=\x_t\sim p_t$ gives
\[
\begin{aligned}
&\mathbb E_{\x_t\sim p_t}
\left[
\sum_{\x\in\X}p_t(\x)
\left\langle
\Phi(\x),A_t\Phi(\x_t)
\right\rangle_{\mathcal H}^2
\;\middle|\;
\mathcal F_{t-1}
\right] \\
&\qquad =
\mathbb E_{\z\sim p_t}
\left[
\left\langle
A_t\Phi(\z),\Sigma_t A_t\Phi(\z)
\right\rangle_{\mathcal H}
\right] \\
&\qquad =
\operatorname{Tr}\!\left(\Sigma_t A_t\Sigma_t A_t\right).
\end{aligned}
\]
Since $A_t=(\Sigma_t+\lambda I_{\mathcal H})^{-1}$ is a spectral function
of $\Sigma_t$, the two operators commute. Hence
\[
\operatorname{Tr}\!\left(\Sigma_t A_t\Sigma_t A_t\right)
=
\operatorname{Tr}\!\left((A_t\Sigma_t)^2\right).
\]
Let $\{\mu_{t,i}\}_{i\ge 1}$ be the eigenvalues of $\Sigma_t$. Then the
eigenvalues of $A_t\Sigma_t$ are
\[
    \frac{\mu_{t,i}}{\mu_{t,i}+\lambda},
    \qquad i\ge 1.
\]
Therefore,
\[
\operatorname{Tr}\!\left((A_t\Sigma_t)^2\right)
=
\sum_{i\ge 1}
\left(\frac{\mu_{t,i}}{\mu_{t,i}+\lambda}\right)^2
\le
\sum_{i\ge 1}
\frac{\mu_{t,i}}{\mu_{t,i}+\lambda}
=
d_{\mathrm{eff}}(\Sigma_t,\lambda),
\]
where the inequality follows because
$\mu_{t,i}/(\mu_{t,i}+\lambda)\in[0,1]$ for every $i$.
This proves the claim.
\end{proof}

%------------------------------------------------------------------
The final lemma of this subsection bounds the effective dimension $d_*(\lambda)$ directly by the standard $T$-point maximum information gain used in the stochastic kernel bandit literature \citep{srinivas2010gp,valko2013finite,vakili2021information}. This is the technical bridge that lets us reformulate our effective-dimension based regret bound in the conventional information gain form.

\begin{restatable}[Effective dimension vs.\ maximum information gain]{myLemma}{effdimvsinfogain}
\label{lem:effdim-vs-infogain}
Assume $\X$ is finite, let $\Phi:\X\to\H$ be a feature map with $\|\Phi(\x)\|_\H^2 \le 1$ for every $\x\in\X$, and define
\[
    \Sigma(\nu) := \E_{\x\sim\nu}[\Phi(\x)\Phi(\x)^\top],
    \qquad
    d_*(\lambda) := \max_{\nu\in\Delta(\X)} \Tr\!\bigl((\Sigma(\nu) + \lambda I_\H)^{-1}\Sigma(\nu)\bigr).
\]
Then, for any $T\ge 1$ and any $\lambda \ge 1/T$, we have
\[
    d_*(\lambda)
    \;\le\;
    2\,\max_{\x_1,\ldots,\x_T\in\X}
    \log\det\!\left(I_\H + \frac{1}{T\lambda}\sum_{t=1}^T \Phi(\x_t)\Phi(\x_t)^\top\right).
\]
\end{restatable}

\begin{proof}
By \citet[Proposition~5]{calandriello2019gaussian}, the effective dimension is bounded by an operator-level log-determinant quantity:
\begin{equation}\label{eq:effdim-cal}
    d_*(\lambda)
    \;\le\;
    \max_{\nu\in\Delta(\X)}\log\det\!\left(I_\H + \frac{\Sigma(\nu)}{\lambda}\right).
\end{equation}
It thus suffices to bound the right-hand side by the $T$-point information gain in the lemma. To this end, set $\alpha := T\lambda$, so that $\alpha\ge 1$ by the assumption $\lambda\ge 1/T$, and $\Sigma(\nu)/\lambda = T\Sigma(\nu)/\alpha$. Fix any $\nu\in\Delta(\X)$, write $M_\nu := T\Sigma(\nu)$, and define $F(A) := \log\det(I_\H + A/\alpha)$ on the positive semidefinite cone. We will construct $T$ points $\x_1,\ldots,\x_T\in\X$ such that
\begin{equation}\label{eq:effdim-goal}
    F\!\left(\sum_{t=1}^T \Phi(\x_t)\Phi(\x_t)^\top\right)
    \;\ge\;
    \tfrac{1}{2}\,F(M_\nu),
\end{equation}
which, combined with \eqref{eq:effdim-cal} and taking the maximum over $\nu\in\Delta(\X)$, yields the lemma.

We construct the points greedily: set $V_0 := 0$ and, for $t=0,1,\ldots,T-1$, choose
\begin{equation}\label{eq:effdim-greedy}
    \x_{t+1} \;\in\; \argmax_{\x\in\X}\|\Phi(\x)\|^2_{(\alpha I_\H + V_t)^{-1}},
    \qquad
    V_{t+1} := V_t + \Phi(\x_{t+1})\Phi(\x_{t+1})^\top,
\end{equation}
where the maximizer exists since $\X$ is finite. With the trace inner product $\langle A,B\rangle_{\mathrm{tr}}=\Tr(AB)$, we have $\nabla F(A) = (\alpha I_\H + A)^{-1}$, and by concavity of $F$,
\begin{align*}
    F(M_\nu) - F(V_t)
    &\le \Tr\!\bigl((\alpha I_\H + V_t)^{-1}(M_\nu - V_t)\bigr) \\
    &\le \Tr\!\bigl((\alpha I_\H + V_t)^{-1} M_\nu\bigr) \\
    &= T\,\E_{\x\sim\nu}\!\bigl[\,\|\Phi(\x)\|^2_{(\alpha I_\H + V_t)^{-1}}\,\bigr] \\
    &\le T\,\|\Phi(\x_{t+1})\|^2_{(\alpha I_\H + V_t)^{-1}},
\end{align*}
where the second inequality drops the non-positive term $-\Tr((\alpha I_\H + V_t)^{-1}V_t)$ (note that $V_t\succeq 0$ and $(\alpha I_\H+V_t)^{-1}\succeq 0$), the equality uses $M_\nu = T\,\E_{\x\sim\nu}[\Phi(\x)\Phi(\x)^\top]$, and the last step is the defining property of $\x_{t+1}$ in \eqref{eq:effdim-greedy}. On the other hand, the matrix determinant lemma gives
\begin{equation}
    F(V_{t+1}) - F(V_t)
    \;=\;
    \log\!\bigl(1 + \|\Phi(\x_{t+1})\|^2_{(\alpha I_\H + V_t)^{-1}}\bigr). \label{eq:F_diff}
\end{equation}
Let $z_t := \|\Phi(\x_{t+1})\|^2_{(\alpha I_\H + V_t)^{-1}}$. Since $V_t\succeq 0$, $\alpha\ge 1$, and $\|\Phi(\x)\|_\H^2 \le 1$, we have $z_t \le \tfrac{1}{\alpha}\|\Phi(\x_{t+1})\|_\H^2 \le 1$. Applying $\log(1+z)\ge (\log 2)\,z$ for $z\in[0,1]$ and combining with \eqref{eq:F_diff} gives
\begin{equation}\label{eq:effdim-step}
    F(M_\nu) - F(V_t)
    \;\le\;
    \frac{T}{\log 2}\,\bigl(F(V_{t+1}) - F(V_t)\bigr).
\end{equation}

We conclude the proof using a telescoping argument. Let $\Delta_t := F(M_\nu) - F(V_t)$. If $\Delta_t \le 0$ for some $t\le T$, then since $F(V_s)$ is non-decreasing in $s$ we have $F(V_T)\ge F(V_t)\ge F(M_\nu)$, and \eqref{eq:effdim-goal} follows. Otherwise $\Delta_t > 0$ for all $t<T$, and \eqref{eq:effdim-step} rearranges to $\Delta_{t+1} = \Delta_t - (F(V_{t+1})-F(V_t)) \le (1 - \tfrac{\log 2}{T})\Delta_t$. Since $V_0 = 0$, we have $\Delta_0 = F(M_\nu)$, and iterating $T$ times yields
\[
    \Delta_T
    \;\le\;
    \Bigl(1 - \tfrac{\log 2}{T}\Bigr)^T F(M_\nu)
    \;\le\;
    e^{-\log 2}\,F(M_\nu)
    \;=\;
    \tfrac{1}{2}\,F(M_\nu),
\]
where the second inequality uses $(1-x)^T\le e^{-xT}$. Therefore $F(V_T)\ge \tfrac{1}{2}\,F(M_\nu)$, which establishes \eqref{eq:effdim-goal} and completes the proof.
\end{proof}

\subsection{Proof of \pref{lem:exp2-learn-bias} (Learner-side Residual)}
\label{proof:exp2-learn-bias}

The following  lemma bounds the learner-side residual terms introduced by the
regularized estimator. It shows that these terms can be controlled by the effective dimension induced by the covariance operator
$\Sigma_t$. 

\learnerresidual*

\begin{proof}
Define the regularization bias operator
\[
    E_t
    :=
    I_\H-(\Sigma_t^\lambda)^{-1}\Sigma_t
    =
    \lambda(\Sigma_t^\lambda)^{-1}.
\]
Since $\Sigma_t^\lambda=\Sigma_t+\lambda I_\H$, we have
$0\preceq E_t\preceq I_\H$.  Moreover, recalling that
\[
    c_t(\x)
    =
    B\sqrt{\lambda}\,
    \norm{\Phi(\x)}_{(\Sigma_t^\lambda)^{-1}}
    =
    B\norm{\Phi(\x)}_{E_t},
\]
we can express both residual terms $B_T^{\x_t}$ and $C_T^{\x_t}$ in terms of the same operator $E_t$.

As for the term $B_T^{\x_t}$,  by definition we have
\[
    B_T^{\x_t}
    =
    \E\!\left[
        \sum_{t=1}^T\sum_{\x\in\X}
        p_t(\x)\,
        \inner{\Phi(\x)}{\w_t-\wht}_\H
    \right].
\]
Conditioning on $\mathcal F_{t-1}$, the only remaining randomness is in
$\x_t$.  Hence, by~\pref{lem:exp2-cond-mean},
\[
    \E[\wht\mid\mathcal F_{t-1}]
    =
    (\Sigma_t^\lambda)^{-1}\Sigma_t\w_t.
\]
Applying the tower property gives
\begin{align*}
    B_T^{\x_t}
    &=
    \E\!\left[
        \sum_{t=1}^T\sum_{\x\in\X}
        p_t(\x)\,
        \inner{\Phi(\x)}
        {\w_t-\E[\wht\mid\mathcal F_{t-1}]}_\H
    \right] \\
    &=
    \E\!\left[
        \sum_{t=1}^T\sum_{\x\in\X}
        p_t(\x)\,
        \inner{\Phi(\x)}
        {\bigl(I_\H-(\Sigma_t^\lambda)^{-1}\Sigma_t\bigr)\w_t}_\H
    \right] \\
    &=
    \E\!\left[
        \sum_{t=1}^T\sum_{\x\in\X}
        p_t(\x)\,
        \inner{\Phi(\x)}{E_t\w_t}_\H
    \right].
\end{align*}
Since $E_t\succeq 0$, Cauchy--Schwarz in the $E_t$-seminorm yields
\[
    \inner{\Phi(\x)}{E_t\w_t}_\H
    =
    \inner{E_t^{1/2}\Phi(\x)}{E_t^{1/2}\w_t}_\H
    \le
    \norm{\Phi(\x)}_{E_t}\norm{\w_t}_{E_t}.
\]
Using $0\preceq E_t\preceq I_\H$ and Assumption~\ref{assum:bounded-kernel}, we have $\norm{\w_t}_{E_t}
    \le
    \norm{\w_t}_\H
    \le
    B.$ 
Therefore, one can further bound the bias term as
\[
    B_T^{\x_t}
    \le
    B\,
    \E\!\left[
        \sum_{t=1}^T\sum_{\x\in\X}
        p_t(\x)\norm{\Phi(\x)}_{E_t}
    \right].
\]
On the other hand, by the definition of $C_T^{\x_t}$, we have
\[
    C_T^{\x_t}
    =
    B\,
    \E\!\left[
        \sum_{t=1}^T\sum_{\x\in\X}
        p_t(\x)\norm{\Phi(\x)}_{E_t}
    \right].
\]
Combining the two displayed equations, we obtain
\begin{equation}
    B_T^{\x_t}+C_T^{\x_t}
    \le
    2B\,
    \E\!\left[
        \sum_{t=1}^T\sum_{\x\in\X}
        p_t(\x)\norm{\Phi(\x)}_{E_t}
    \right].
    \label{eq:learner-residual-common-term}
\end{equation}

It remains to relate the right-hand side to the effective dimension.  For
any $\alpha>0$, Young's inequality gives
\[
    B\norm{\Phi(\x)}_{E_t}
    \le
    \alpha\norm{\Phi(\x)}_{E_t}^2
    +
    \frac{B^2}{\alpha}.
\]
Multiplying by $p_t(\x)$ and summing over $\x\in\X$, we get
\[
    \sum_{\x\in\X}p_t(\x)\,
    B\norm{\Phi(\x)}_{E_t}
    \le
    \alpha
    \sum_{\x\in\X}p_t(\x)\norm{\Phi(\x)}_{E_t}^2
    +
    \frac{B^2}{\alpha},
\]
where we used $\sum_{\x\in\X}p_t(\x)=1$.  Substituting this into
\eqref{eq:learner-residual-common-term} gives
\begin{align*}
    B_T^{\x_t}+C_T^{\x_t}
    &\le
    2\,
    \E\!\left[
        \sum_{t=1}^T
        \left(
            \alpha
            \sum_{\x\in\X}p_t(\x)\norm{\Phi(\x)}_{E_t}^2
            +
            \frac{B^2}{\alpha}
        \right)
    \right] \\
    &=
    2\alpha\,
    \E\!\left[
        \sum_{t=1}^T
        \sum_{\x\in\X}p_t(\x)\norm{\Phi(\x)}_{E_t}^2
    \right]
    +
    \frac{2B^2T}{\alpha}.
\end{align*}

We now handle the quadratic term.  By the definition $\norm{\Phi(\x)}_{E_t}^2
    =
    \inner{\Phi(\x)}{E_t\Phi(\x)}_\H,$
we have
\begin{align*}
    \sum_{\x\in\X}p_t(\x)\norm{\Phi(\x)}_{E_t}^2
    &=
    \sum_{\x\in\X}p_t(\x)
    \inner{\Phi(\x)}{E_t\Phi(\x)}_\H \\
    &=
    \Tr\!\left(
        E_t
        \sum_{\x\in\X}p_t(\x)\Phi(\x)\Phi(\x)^\top
    \right) \\
    &=
    \Tr(E_t\Sigma_t).
\end{align*}
Using $E_t=\lambda(\Sigma_t^\lambda)^{-1}$, we obtain
\[
    \Tr(E_t\Sigma_t)
    =
    \lambda\,
    \Tr\bigl((\Sigma_t^\lambda)^{-1}\Sigma_t\bigr).
\]
Therefore,
\[
    \sum_{\x\in\X}p_t(\x)\norm{\Phi(\x)}_{E_t}^2
    =
    \lambda\,
    \Tr\bigl((\Sigma_t^\lambda)^{-1}\Sigma_t\bigr).
\]
Substituting this identity into the previous bound yields
\[
    B_T^{\x_t}+C_T^{\x_t}
    \le
    2\alpha\lambda\,
    \E\left[
        \sum_{t=1}^T
        \Tr\bigl((\Sigma_t^\lambda)^{-1}\Sigma_t\bigr)
    \right]
    +
    \frac{2B^2T}{\alpha} = 2\alpha \lambda\cdot \E\left[\sum_{t=1}^T d_{\mathrm{eff}}(\Sigma_t,\lambda)\right] + \frac{2B^2 T}{\alpha},
\]
which proves the claim by choosing $\alpha = B\sqrt{\frac{T}{\lambda\,\E\!\left[\sum_{t=1}^T d_{\mathrm{eff}}(\Sigma_t,\lambda)\right]}}.$ Note that If $\mathbb{E}\left[\sum_t d_{\mathrm{eff}}(\Sigma_t,\lambda)\right]=0$, the claim is trivial.
\end{proof}

\subsection{Proof of \pref{lem:exp2-comp-bias} (Comparator-side Residual)}
\label{proof:exp2-comp-bias}

The following lemma shows that the comparator-side residual is non-positive.
For any fixed comparator $\x_*$, the correction term dominates the bias induced
by regularization in the comparator direction. Consequently, this residual does
not contribute positively to the regret. This is a key structural property of
the analysis, as it eliminates the need to further bound the comparator-side
bias and leaves only the learner-side residual to be controlled.

\comparatorresidual*

\begin{proof}
Define the regularization bias operator
\[
    E_t
    :=
    I_\H-(\Sigma_t^\lambda)^{-1}\Sigma_t
    =
    \lambda(\Sigma_t^\lambda)^{-1}.
\]
As in the proof of Lemma~\ref{lem:exp2-learn-bias}, we have $0 \preceq E_t \preceq I_\H$. By \pref{lem:exp2-cond-mean}, conditioning on $\mathcal F_{t-1}$, we have
\[
\E[\wht \mid \mathcal F_{t-1}]
=
(\Sigma_t^\lambda)^{-1}\Sigma_t \w_t,
\]
and hence
\[
\E[\w_t - \wht \mid \mathcal F_{t-1}]
=
E_t \w_t.
\]
Since the comparator $\x_*$ is fixed, $\Phi(\x_*)$ is $\mathcal F_{t-1}$-measurable. Applying the tower property,
\begin{align*}
B_T^{\x_*}
&=
\E\!\left[\sum_{t=1}^T
\inner{\Phi(\x_*)}{\w_t - \wht}_\H\right] \\
&=
\E\!\left[\sum_{t=1}^T
\inner{\Phi(\x_*)}{E_t \w_t}_\H\right].
\end{align*}
Recall that
\[
C_T^{\x_*}
= B \sqrt{\lambda} \E\left[\sum_{t=1}^T \norm{\Phi(\x_*)}_{(\SigLam)^{-1}}\right]= 
B\,\E\!\left[\sum_{t=1}^T \|\Phi(\x_*)\|_{E_t}\right].
\]
We now combine $B_T^{\x_*}$ and $C_T^{\x_*}$:
\begin{align*}
- B_T^{\x_*} - C_T^{\x_*}
&=
-\E\!\left[\sum_{t=1}^T
\inner{\Phi(\x_*)}{E_t \w_t}_\H\right]
-
B\,\E\!\left[\sum_{t=1}^T \|\Phi(\x_*)\|_{E_t}\right].
\end{align*}
Since $E_t \succeq 0$, Cauchy--Schwarz in the $E_t$-seminorm yields
\[
|\inner{\Phi(\x_*)}{E_t \w_t}_\H|
=
\vert\inner{E_t^{1/2}\Phi(\x_*)}{E_t^{1/2}\w_t}_\H\vert
\le
\|\Phi(\x_*)\|_{E_t}\,\|\w_t\|_{E_t}.
\]
Therefore,
\begin{align*}
- B_T^{\x_*} - C_T^{\x_*}
&\le
\E\!\left[
\sum_{t=1}^T
\left(
\|\Phi(\x_*)\|_{E_t}\,\|\w_t\|_{E_t}
-
B\,\|\Phi(\x_*)\|_{E_t}
\right)
\right]\\
&=
\E\!\left[
\sum_{t=1}^T
\|\Phi(\x_*)\|_{E_t}
\bigl(
\|\w_t\|_{E_t} - B
\bigr)
\right]\\
&\le 0,
\end{align*}
where the last inequality holds due to $\|\w_t\|_{E_t} \le \|\w_t\|_\H \leq B$ by Assumption~\ref{assum:bounded-kernel}.

\end{proof}

%------------------------------------------------------------------
\subsection{Proof of \pref{lem:exp2-reg-ew} (Regret Bound via Exponential Weights)}
\label{proof:exp2-reg-ew}

The following lemma bounds the estimated regret term arising from the
exponential-weights update. Under a suitable exploration condition, the proxy
losses are uniformly bounded, allowing a standard exponential-weights analysis.
The resulting bound separates the contribution of the effective dimension term, the entropy term $\log|\X|/\eta$, and the
cost of explicit exploration.

\estimatedregret*
\begin{proof}
We can decompose $\Reg_T^{\hat\ell}$ into two terms using $p_t = (1-\gamma)q_t + \gamma \nu_G$:
\begin{align*}
    \Reg_T^{\hat\ell}
    \;=\;
    &(1-\gamma)\underbrace{
        \E\!\left[\sum_{t=1}^T \sum_{\x\in\X}\bigl(q_t(\x)-p_*(\x)\bigr)\hat{\ell}_t(\x)\right]
    }_{\texttt{Term~(A)}}
    \\
    &\quad+\;
    \gamma\underbrace{
        \E\!\left[\sum_{t=1}^T \sum_{\x\in\X}\bigl({\nu}_G(\x)-p_*(\x)\bigr)\hat{\ell}_t(\x)\right]
    }_{\texttt{Term~(B)}}.
\end{align*}

\paragraph{Bounding Term (A).} When $\gamma=1$, the coefficient $1-\gamma$ makes Term~(A) vanish and the regret reduces to $\gamma \cdot \texttt{Term~(B)}$. It therefore suffices to consider $\gamma\in(0,1)$ in the following. Term (A) is the regret of the distribution $q_t$ generated by the exponential weights algorithm. According to Lemma~\ref{lem:exp2-bounded}, we have $\vert \eta \hat{\ell}_t(\x)\vert \leq 1$. Then, the standard analysis for the exponential weights algorithm (e.g., see~\citep[Theorem 1]{bubeck2012towards}) yields
\begin{align}
    \texttt{Term~(A)}
    \;\le\;&
    \eta\,\E\!\left[\sum_{t=1}^T \sum_{\x\in\X} q_t(\x)\hat{\ell}_t(\x)^2\right]
    + \frac{\log|\X|}{\eta}
    \notag\\
    \;\le\;&
    \frac{\eta}{1-\gamma}\,
    \E\!\left[\sum_{t=1}^T \sum_{\x\in\X} p_t(\x)\hat{\ell}_t(\x)^2\right]
    + \frac{\log|\X|}{\eta}\notag\\
    \;=\;& \frac{\eta}{1-\gamma} \E\left[ \sum_{t=1}^T \E_{\x_t\sim p_t}\Big[\sum_{\x\in\X} p_t(\x) \hat{\ell}_t(\x)^2\ \vert\ \F_{t-1}\Big] + \frac{\log \vert\X\vert}{\eta}\right]
    \label{eq:exp2-termA}
\end{align}
where the second inequality uses $p_t(\x)=(1-\gamma)q_t(\x)+\gamma\nu_G(\x)\ge (1-\gamma)q_t(\x)$. 

We can further upper bound the quadratic term on the right-hand side by
\begin{align}
    &\E_{\x_t\sim p_t}\!\left[
        \sum_{\x\in\X} p_t(\x)\hat{\ell}_t(\x)^2
        \,\middle|\,\mathcal{F}_{t-1}
    \right]\notag\\
    &\le\;
    2\,\E_{\x_t\sim p_t}\!\left[
        \sum_{\x\in\X} p_t(\x)\,
        \vert\inner{\Phi(\x_t)}{\w_t}_\H\vert^2 \cdot
        \inner{\Phi(\x)}{(\Sigma_t^{\lambda})^{-1}\Phi(\x_t)}_\H^2
        \,\middle|\,\mathcal{F}_{t-1}
    \right]
    + 2\sum_{\x\in\X}p_t(\x)c_t(\x)^2\notag\\
    &\le\;
    2B^2\,\E_{\x_t\sim p_t}\!\left[
        \sum_{\x\in\X} p_t(\x)\,
        \inner{\Phi(\x)}{(\Sigma_t^{\lambda})^{-1}\Phi(\x_t)}_\H^2
        \,\middle|\,\mathcal{F}_{t-1}
    \right]
    + 2\sum_{\x\in\X}p_t(\x)c_t(\x)^2\notag\\
        & \leq\; 2B^2  d_{\mathrm{eff}}(\Sigma_t,\lambda)+ 2\sum_{\x\in\X}p_t(\x)c_t(\x)^2,
    \label{eq: exp2_variance_term}
\end{align}
where:
\begin{itemize}[leftmargin=4ex]
    \item the first inequality follows from the definition
    $\hat{\ell}_t(\x) = \inner{\Phi(\x_t)}{\w_t}_\H \inner{\Phi(\x)}{(\Sigma_t^{\lambda})^{-1}\Phi(\x_t)}_\H - c_t(\x)$
    and the fact that $(u-v)^2 \le 2 u^2 + 2 v^2$ with $u = \inner{\Phi(\x_t)}{\w_t}_\H \inner{\Phi(\x)}{(\Sigma_t^{\lambda})^{-1}\Phi(\x_t)}_\H$ and $v = c_t(\x)$;
    \item the second inequality uses \pref{assum:bounded-kernel};
    \item the last step applies \pref{lem:quadratic-leverage}.
\end{itemize}
For the second term in \eqref{eq: exp2_variance_term}, using the definition of \(c_t(\x)\), we have
\begin{align}
    \sum_{\x\in\X}p_t(\x)c_t(\x)^2
    \;=\;&
    B^2\lambda\,
    \mathrm{Tr}\!\left((\Sigma_t^{\lambda})^{-1}\Sigma_t\right) = B^2\lambda d_{\mathrm{eff}}(\Sigma_t,\lambda).
     \label{eq: exp2_variance_term_term_2}
\end{align}

Substituting \eqref{eq: exp2_variance_term} and \eqref{eq: exp2_variance_term_term_2} into \eqref{eq:exp2-termA} gives
\begin{align*}
    (1-\gamma)\cdot \texttt{Term~(A)}
    \;\le\;&
    2(1+\lambda)B^2\eta\cdot \sum_{t=1}^T \E\left[d_{\mathrm{eff}}(\Sigma_t,\lambda)\right] +
    \frac{\log|\X|}{\eta}.
\end{align*}

\paragraph{Bounding Term (B).} 
Using the tower property together with \pref{lem:exp2-cond-mean}, we have
\[
\texttt{Term~(B)}
=
\E \Bigg[
    \sum_{t=1}^T \sum_{\x\in\X} (\nu_G(\x)-p_*(\x))
    \Bigl(
        \inner{\Phi(\x)}{(\Sigma_t^\lambda)^{-1}\Sigma_t\,\w_t}_\H
        - c_t(\x)
    \Bigr)
\Bigg].
\]

We bound the two contributions separately. For the first part, since $(\Sigma_t^\lambda)^{-1}\Sigma_t \preceq I_\H$ and $\|\w_t\|_\H \le B$, we have
\[
\left|\inner{\Phi(\x)}{(\Sigma_t^\lambda)^{-1}\Sigma_t\,\w_t}_\H\right|
\le \|\Phi(\x)\|_\H\,\|(\Sigma_t^\lambda)^{-1}\Sigma_t\,\w_t\|_\H
\le B
\quad \text{for all $\x\in\X$}.
\]
Moreover, using $\sum_{\x\in\X} |\nu_G(\x)-p_*(\x)| \le 2$ gives
\[
\left|\sum_{\x\in\X} (\nu_G(\x)-p_*(\x)) \inner{\Phi(\x)}{(\Sigma_t^\lambda)^{-1}\Sigma_t\,\w_t}_\H\right|
\le 2B.
\]

For the correction term,
\[
-\sum_{\x\in\X} (\nu_G(\x)-p_*(\x)) c_t(\x)
= c_t(\x_*) - \sum_{\x\in\X} \nu_G(\x)\, c_t(\x) \le c_t(\x_*),
\]
since $\sum_{\x\in\X} \nu_G(\x)\, c_t(\x) \ge 0$.

Combining both contributions and summing over $t=1,\dots,T$ and taking expectation yields
\[
\texttt{Term~(B)} \le 2B T + C_T^{\x_*}.
\]

We can further upper bound the term $C_T^{\x_*}$ by 
\begin{align*}
    C_T^{\x_*} = B \E\Big[\sum_{t=1}^T \|\Phi(\x_*)\|_{E_t}\Big]   \leq B \E\left[\sqrt{T\sum_{t=1}^T \norm{\Phi(\x_*)}^2_{E_t}}\right],
\end{align*}
where $E_t = \lambda (\Sigma_t^\lambda)^{-1}$ and the last step is due to the Cauchy–Schwarz inequality.

Now, writing $\|\Phi(\x_*)\|_{E_t}^2 = \lambda \|\Phi(\x_*)\|_{(\Sigma_t^\lambda)^{-1}}^2$, and using $\Sigma_t \succeq \gamma \Sigma(\nu_G)$, we have
\[
(\Sigma_t^\lambda)^{-1} \preceq \frac{1}{\gamma} \Big(\Sigma(\nu_G) + \frac{\lambda}{\gamma} I_\H \Big)^{-1}.
\]
Hence,
\[
\|\Phi(\x_*)\|_{E_t}^2
\le \frac{\lambda}{\gamma} \Tr\Big[ \big(\Sigma(
    \nu_D) + \frac{\lambda}{\gamma} I_\H\big)^{-1} \Sigma(\nu_D) \Big]
= \frac{\lambda}{\gamma}d_{\mathrm{eff}}(\Sigma(\nu_D), \lambda/\gamma),
\]
where the inequality holds following the same arguments in deriving~\eqref{eq:max-Phi} in the proof of Lemma~\ref{lem:exp2-bounded}.

Substituting back, we obtain
\[
C_T^{\x_*} \le B\sqrt{\frac{\lambda T}{\gamma}\sum_{t=1}^T d_{\mathrm{eff}}(\Sigma(\nu_D),\lambda/\gamma)}
\]

\paragraph{Conclusion.} 
Combining the bounds for $(1-\gamma)\cdot \texttt{Term~(A)}$ and $\gamma \cdot \texttt{Term~(B)}$ gives \eqref{eq:exp2-reg-ew-bound-effective}.
\end{proof}

\subsection{Proof of \pref{cor: maincor} and Applications to Specific Kernels}
\label{app:cor1}

Here we provide the omitted proof details for \pref{cor: maincor} and the subsequent applications to specific kernels. 

\begin{proof}[Proof of Corollary \ref{cor: maincor}]
Recall that \pref{thm: main} bounds the regret in terms of
\[
    d_*(\rho)
    =
    \max_{\nu\in\Delta(\mathcal X)}
    \operatorname{Tr}
    \left(
        (\Sigma(\nu)+\rho I_{\mathcal H})^{-1}
        \Sigma(\nu)
    \right)~,
\]
where
$
    \Sigma(\nu)
    =
    \mathbb E_{\x\sim\nu}
    \left[
        \Phi(x)\Phi(x)^\top
    \right]~.
$ Fixing $\alpha = 1$, we have
$
    \lambda=1/T
$ and \pref{lem:effdim-vs-infogain} gives
\[
     d_*(\lambda) \leq 2\max_{\x_1,\ldots,\x_T\in\mathcal X}
    \log\det\left(
        I+
        \sum_{t=1}^T
        \Phi(x_t)\Phi(x_t)^\top
    \right) =:
    4\bar{\gamma}_T.
\]
We will also need to control $d_*(\lambda/\gamma)$, where
$\gamma$ denotes the mixing coefficient in \pref{alg:exp2}. Since 
$\gamma \in (0,1]$, we have
\[
    \frac{\lambda}{\gamma}
    =
    \frac{1}{\gamma T}
    \ge
    \frac{1}{T}.
\]
Moreover, since $d_*(\rho)$ is non-increasing in $\rho$, we have
\begin{equation}
    d_*(\lambda/\gamma)
    \le
    d_*(\lambda)
    \le
    4\bar{\gamma}_T.
    \label{eq: D_T}
\end{equation}
We choose
\begin{equation}
    \eta
    =
    \frac{1}{2B}
    \sqrt{
        \frac{\log(e|\mathcal X|)}{2(1+\lambda)d_*(\lambda)T}
    },
    \qquad
    \gamma
    = \min \left\{  
    \sqrt{
        \frac{2d_*(\lambda) \log(e|\mathcal X|)}{(1+\lambda)T}
    },1\right\}.
    \label{eq: learning_rate_m}
\end{equation}
We claim that it suffices to assume that $T \geq 2d_*(\lambda) \log(e|\mathcal X|)$; this is due to the fact that when $T \leq  2d_*(\lambda) \log(e|\mathcal X|) $, we immediately have $\Reg_T \leq 2BT \leq 2B \sqrt{ 2Td_*(\lambda) \log(e|\mathcal X|)}$. 

Under the condition $T \geq 2d_*(\lambda) \log(e|\mathcal X|)$ we have $\gamma < 1$, and recalling that $\lambda = 1/T$, a direct substitution readily verifies that $\lambda\le 1/(2\eta B)$.  
Moreover, we observe from \eqref{eq: D_T} that
\[
    d_*(\lambda/\gamma)
    \le
    2 d_*(\lambda)
    =
    \frac{\gamma}{2\eta B}.
\]
Hence, both conditions of \pref{thm: main} are satisfied.  
Applying \pref{thm: main} (with
$\lambda=1/T$), we obtain
\[
\begin{aligned}
    \Reg_T
    &\le
    \frac{\log|\mathcal X|}{\eta}
    +
    2\Big(1+\frac{1}{T}\Big)B^2\eta d_*(\lambda) T 
    +
    4B\sqrt{ T d_*(\lambda)}      
    +
    B\sqrt{ \gamma T  d_*(\lambda)}
    +
    2\gamma BT.
\end{aligned}
\]
The first two terms satisfy
\[
    \frac{\log|\mathcal X|}{\eta}
    +
    2\Big(1+\frac{1}{T}\Big)B^2\eta d_*(\lambda)T
    \le
    4B\sqrt{(1+\lambda) d_*(\lambda) T \log(e|\mathcal X|)},
\]
and the fourth term satisfies the following since $\gamma\le1$:
\[
     B\sqrt{ \gamma T  d_*(\lambda)}
    \le
    B\sqrt{Td_*(\lambda)}.
\]
For the final term, we have
\[
    2\gamma BT
    =
    2BT
    \sqrt{
        \frac{2d_*(\lambda) \log(e|\mathcal X|) }{(1+\lambda)T}
    }
    \le
    2B\sqrt{2d_*(\lambda) T \log(e|\mathcal X|) }~.
\]
Combining these bounds and using \eqref{eq: D_T} gives
\begin{equation}
    \Reg_T
    =
    \mathcal{O}
    \left(
        B\sqrt{d_*(\lambda) T\log(e|\mathcal X|)}
    \right)
    =
    \mathcal{O}
    \left(
        B\sqrt{\bar{\gamma}_T T\log(|\mathcal X|)}
    \right).
    \label{eq: regret_m}
\end{equation}
\end{proof}

\textit{Application to specific kernels.} 
For the Matérn$(\nu,d)$ kernel, the standard
information-gain bound is given by \citep{vakili2021information}
\[
    \bar{\gamma}_T
    =
    \widetilde{\mathcal{O}}
    \left(
        T^{d/(2\nu+d)}
    \right).
\]
Substituting this into \eqref{eq: regret_m}, we obtain
\[
\begin{aligned}
    R_T
    &=
    \widetilde{\mathcal{O}}
    \left(
        B
        \sqrt{
            T\cdot T^{d/(2\nu+d)}
            \log(e|\mathcal X|)
        }
    \right)                                                           \\
    &=
    \widetilde{\mathcal{O}}
    \left(
        B
        T^{\frac12+\frac{d}{2(2\nu+d)}}
        \sqrt{\log(e|\mathcal X|)}
    \right)                                                           \\
    &=
    \widetilde{\mathcal{O}}
    \left(
        B
        T^{(\nu+d)/(2\nu+d)}
        \sqrt{\log(e|\mathcal X|)}
    \right).
\end{aligned}
\]
In particular, if $\log|\mathcal X|=\mathcal{O}(\log T)$ and $B$ is treated as a constant, then
\[
    \Reg_T
    =
    \mathcal{O}\left(
        T^{(\nu+d)/(2\nu+d)}
    \right),
\]
which proves the desired claim.

For the Squared Exponential kernel, the standard
information-gain bound is given by \citep{vakili2021information}
\[
    \bar{\gamma}_T
    =
    \mathcal{O}\big((\log T)^{d+1}\big)~.
\]
 Substituting this into \eqref{eq: regret_m}, we obtain
\[
\begin{aligned}
    \Reg_T
    &=
    \mathcal{O}\left(
        B
        \sqrt{
            T\log(e|\mathcal X|)
            (\log T)^{d+1}
        }
    \right)                                                     \\
    &=
    \mathcal{O}\left(
        B
        \sqrt{T\log(e|\mathcal X|)}
        (\log T)^{(d+1)/2}
    \right).
\end{aligned}
\]
In particular, if $\log|\mathcal X|=\mathcal{O}(\log T)$ and $B$ is treated as a constant, then
\[
    \Reg_T
    =
    \mathcal{O}\left(
        \sqrt T(\log T)^{(d+2)/2}
    \right)
    =
    \widetilde {\mathcal{O}}(\sqrt T),
\]
which proves the desired claim.

\subsection{Proofs of Corollary~\ref{cor: main} and Corollary~\ref{cor: main_2} (Polynomial and Exponential Eigendecay)}
\label{app:decay}

\begin{proof}[Proof of Corollary~\ref{cor: main}]

Suppose $\ke$ has $(C, \beta)$-polynomial eigenvalue decay defined in \pref{def:decay_1}, \pref{lem:effective-dim} shows that
the effective dimension can be upper bounded by 
\[
    d_*(\lambda)
    \le
    m+\frac{C m^{1-\beta}}{(\beta-1)\lambda}
\]
and
\[
    d_*(\lambda/\gamma)
    \le
    m+\frac{\gamma C m^{1-\beta}}{(\beta-1)\lambda}.
\]
Then, Theorem~\ref{thm: main} gives
\begin{align*}
\Reg_T
\le{}&
2\gamma BT
+
\frac{\log|\X|}{\eta}
+
2(1+\lambda)B^2\eta T
\left(
    m+\frac{C m^{1-\beta}}{(\beta-1)\lambda}
\right) \\
&\quad
+
BT
\sqrt{
    \lambda\gamma m
    +
    \frac{\gamma^2 C m^{1-\beta}}{\beta-1}
}
+
4BT
\sqrt{
    \lambda m
    +
    \frac{C m^{1-\beta}}{\beta-1}
}.
\end{align*}

We now choose
\begin{equation}
\label{eq:parameter-tuning-poly}
    m := \left\lceil T^{1/\beta}\right\rceil,
    \qquad
    \lambda := m^{-\beta},
    \qquad
    \gamma := m^{-\frac{\beta-1}{2}},
    \qquad
    \eta :=
    \frac{1}{2B(1+C/(\beta-1))}
    m^{-\frac{\beta+1}{2}} .
\end{equation}
By this choice,
\[
    m+\frac{C m^{1-\beta}}{(\beta-1)\lambda}
    =
    \left(1+\frac{C}{\beta-1}\right)m
\]
and
\[
    m+\frac{\gamma C m^{1-\beta}}{(\beta-1)\lambda}
    =
    \left(1+\frac{C\gamma}{\beta-1}\right)m
    \le
    \left(1+\frac{C}{\beta-1}\right)m,
\]
where we used $\gamma\le 1$. Substituting these bounds into the regret bound yields
\begin{align*}
\Reg_T
\le{}&
2\gamma BT
+
\frac{\log|\X|}{\eta}
+
2(1+\lambda)B^2\eta T
\left(1+\frac{C}{\beta-1}\right)m \\
&\quad
+
BT
\sqrt{
    \lambda\gamma m
    \left(1+\frac{C\gamma}{\beta-1}\right)
}
+
4BT
\sqrt{
    \lambda m
    \left(1+\frac{C}{\beta-1}\right)
}.
\end{align*}

Since $m=\lceil T^{1/\beta}\rceil$, we have
$m\le T^{1/\beta}+1$ and $m^{-a}\le T^{-a/\beta}$ for any $a>0$.
Therefore,
\[
\begin{aligned}
\Reg_T
\le{}&
2B\left(1+\frac{C}{\beta-1}\right)\log|\X|\,
\left(T^{1/\beta}+1\right)^{\frac{\beta+1}{2}}  \\
&\quad+
B T^{\frac{\beta+1}{2\beta}}
\left(
    3+T^{-1}
    +4\sqrt{1+\frac{C}{\beta-1}}
\right) \\
&\quad+
B T^{\frac{\beta+3}{4\beta}}
\sqrt{
    1+\frac{C T^{-\frac{\beta-1}{2\beta}}}{\beta-1}
}.
\end{aligned}
\]
Since the last term is lower order for $\beta>1$, we obtain
\[
\Reg_T
=
\mathcal{O}\!\left(
B\left(
1+
\left(1+\frac{C}{\beta-1}\right)\log|\X|
+
\sqrt{1+\frac{C}{\beta-1}}
\right)
T^{\frac{\beta+1}{2\beta}}
\right).
\]

It remains to verify the conditions of Theorem~\ref{thm: main}. By the parameter choice,
\[
d_{\mathrm{eff}}(\Sigma(\nu_D),\lambda/\gamma)
\le
m+\frac{\gamma C m^{1-\beta}}{(\beta-1)\lambda}
=
\left(1+\frac{C\gamma}{\beta-1}\right)m
\le
\left(1+\frac{C}{\beta-1}\right)m
=
\frac{\gamma}{2\eta B}.
\]
Moreover,
\[
\lambda=m^{-\beta}
\le 1
\le
\left(1+\frac{C}{\beta-1}\right)m^{\frac{\beta+1}{2}}
=
\frac{1}{2\eta B}.
\]
Thus both conditions of Theorem~\ref{thm: main} are satisfied.
\end{proof}

\begin{proof}[Proof of Corollary~\ref{cor: main_2}]
Suppose $\ke$ has $(C, \beta)$-exponential eigenvalue decay defined in \pref{def:decay_1}, \pref{lem:effective-dim} shows that
the effective dimension can be upper bounded by
\[
    d_{*}(\lambda)
    \le
    m+\frac{C e^{-\beta m}}{(e^\beta-1)\lambda},  \qquad
    d_{*}(\lambda/\gamma)
    \le
    m+\frac{\gamma C e^{-\beta m}}{(e^\beta-1)\lambda}.
\]
Then, Theorem~\ref{thm: main} gives
\begin{align*}
\Reg_T
\le{}&
2\gamma BT
+
\frac{\log|\X|}{\eta}
+
2(1+\lambda)B^2\eta T
\left(
    m+\frac{C e^{-\beta m}}{(e^\beta-1)\lambda}
\right) \\
&\quad
+
BT
\sqrt{
    \lambda\gamma m
    +
    \frac{\gamma^2 C e^{-\beta m}}{e^\beta-1}
}
+
4BT
\sqrt{
    \lambda m
    +
    \frac{C e^{-\beta m}}{e^\beta-1}
}.
\end{align*}

We now choose
\begin{equation}
\label{eq:parameter-tuning-exp}
    m := \left\lceil \frac{\log T}{\beta}\right\rceil,
    \qquad
    \lambda := e^{-\beta m},
\end{equation}
and
\[
    \gamma
    :=
    \sqrt{
        \frac{
            \left(m+\frac{C}{e^\beta-1}\right)\log(e|\X|)
        }{T}
    },
    \qquad
    \eta
    :=
    \frac{\gamma}{
        2B\left(m+\frac{C}{e^\beta-1}\right)
    }.
\]
Since $m=\lceil(\log T)/\beta\rceil$, we have
\[
    \lambda=e^{-\beta m}\le \frac{1}{T}.
\]
Moreover,
\[
    m+\frac{C e^{-\beta m}}{(e^\beta-1)\lambda}
    =
    m+\frac{C}{e^\beta-1},
\]
and
\[
    m+\frac{\gamma C e^{-\beta m}}{(e^\beta-1)\lambda}
    =
    m+\frac{\gamma C}{e^\beta-1}
    \le
    m+\frac{C}{e^\beta-1},
\]
where the last inequality uses $\gamma\le 1$.

Substituting these bounds into the regret bound yields
\begin{align*}
\Reg_T
\le{}&
2\gamma BT
+
\frac{\log|\X|}{\eta}
+
2(1+\lambda)B^2\eta T
\left(
    m+\frac{C}{e^\beta-1}
\right) \\
&\quad
+
BT
\sqrt{
    \lambda\gamma
    \left(
        m+\frac{\gamma C}{e^\beta-1}
    \right)
}
+
4BT
\sqrt{
    \lambda
    \left(
        m+\frac{C}{e^\beta-1}
    \right)
}.
\end{align*}

We estimate the terms separately. First, by the choice of $\gamma$,
\[
    2\gamma BT
    =
    2B
    \sqrt{
        T
        \left(
            m+\frac{C}{e^\beta-1}
        \right)
        \log(e|\X|)
    }.
\]
For the entropy term, since $\log|\X|\le \log(e|\X|)$,
\[
\frac{\log|\X|}{\eta}
=
\frac{
    2B
    \left(
        m+\frac{C}{e^\beta-1}
    \right)
    \log|\X|
}{\gamma}
\le
2B
\sqrt{
    T
    \left(
        m+\frac{C}{e^\beta-1}
    \right)
    \log(e|\X|)
}.
\]
For the variance term, using $\lambda\le 1$,
\[
\begin{aligned}
2(1+\lambda)B^2\eta T
\left(
    m+\frac{C}{e^\beta-1}
\right)
&\le
4B^2\eta T
\left(
    m+\frac{C}{e^\beta-1}
\right) \\
&=
2B\gamma T \\
&=
2B
\sqrt{
    T
    \left(
        m+\frac{C}{e^\beta-1}
    \right)
    \log(e|\X|)
}.
\end{aligned}
\]
For the first square-root term, using $\lambda\le 1/T$ and $\gamma\le 1$,
\[
\begin{aligned}
BT
\sqrt{
    \lambda\gamma
    \left(
        m+\frac{\gamma C}{e^\beta-1}
    \right)
}
&\le
BT
\sqrt{
    \frac{\gamma}{T}
    \left(
        m+\frac{C}{e^\beta-1}
    \right)
} \\
&=
B
\sqrt{
    T\gamma
    \left(
        m+\frac{C}{e^\beta-1}
    \right)
} \\
&\le
B
\sqrt{
    T
    \left(
        m+\frac{C}{e^\beta-1}
    \right)
    \log(e|\X|)
},
\end{aligned}
\]
where the last inequality uses $\gamma\le 1\le \log(e|\X|)$. Similarly,
\[
\begin{aligned}
4BT
\sqrt{
    \lambda
    \left(
        m+\frac{C}{e^\beta-1}
    \right)
}
&\le
4BT
\sqrt{
    \frac{1}{T}
    \left(
        m+\frac{C}{e^\beta-1}
    \right)
} \\
&=
4B
\sqrt{
    T
    \left(
        m+\frac{C}{e^\beta-1}
    \right)
} \\
&\le
4B
\sqrt{
    T
    \left(
        m+\frac{C}{e^\beta-1}
    \right)
    \log(e|\X|)
}.
\end{aligned}
\]
Combining the above estimates gives
\[
\Reg_T
=
\mathcal{O}\!\left(
B
\sqrt{
    T
    \left(
        m+\frac{C}{e^\beta-1}
    \right)
    \log(e|\X|)
}
\right).
\]
Since
$
    m=\left\lceil \frac{\log T}{\beta}\right\rceil,
$
we obtain
\[
\Reg_T
=
\mathcal{O}\!\left(
B
\sqrt{
    T
    \left(
        \frac{\log T}{\beta}
        +
        \frac{C}{e^\beta-1}
    \right)
    \log(e|\X|)
}
\right).
\]
In particular, when $(B,\beta,C)$ are constant and $\log|\X|=O(\log T)$,
we have
$
    \Reg_T = O\!\left(\sqrt{T}\log T\right).
$

It remains to verify the conditions of Theorem~\ref{thm: main}. By the parameter choice,
\[
d_{\mathrm{eff}}(\Sigma(\nu_D),\lambda/\gamma)
\le
m+\frac{\gamma C}{e^\beta-1}
\le
m+\frac{C}{e^\beta-1}
=
\frac{\gamma}{2\eta B}.
\]
Moreover,
\[
\lambda=e^{-\beta m}
\le 1
\le
\frac{
    m+\frac{C}{e^\beta-1}
}{\gamma}
=
\frac{1}{2\eta B},
\]
where we used $\gamma\le 1$. Thus both conditions of Theorem~\ref{thm: main}
are satisfied.
\end{proof}

\section{Further Discussion on Existing and Concurrent Work} \label{app:discussion}

\subsection{Lower Bound from \cite{chatterji2019online}}
\label{app:discussion_1}

In \cite[Theorem 43]{chatterji2019online}, a lower bound of $\tilde{\Omega}(T^{\frac{\beta+1}{2\beta}})$ is given for adversarial kernel bandits in a setting where the kernel has eigenvalues decaying as $O(j^{-\beta})$.  While our upper bound also exhibits $T^{\frac{\beta+1}{2\beta}}$ dependence, there are important differences in the underlying assumptions that makes these bounds incomparable and/or introduces a gap between them.

Perhaps most importantly, the kernel used in \cite[Theorem 43]{chatterji2019online} is an infinite-dimensional linear kernel, as opposed to a kernel with low-dimensional input such as $[0,1]^d$.  Their proof can be adapted to a finite action set of the form $\{-1,1\}^M$ for suitably-chosen $M$ (with all values beyond $M$ implicitly set to zero), but this requires $M$ to grow at rate $T^{\frac{1}{2\beta}}$, thus inducing an action set $|\X|$ satisfying $\log|\X| = \Omega( T^{\frac{1}{2\beta}} )$, rather than the usual $\log|\X| = O(\log T)$.  Accordingly, the $\log|\X|$ dependence in our upper bound (and in related upper bounds, including that of \cite{chatterji2019online}) would be significant due to being polynomial in $T$, and the overall regret scaling would be strictly higher than $T^{\frac{\beta+1}{2\beta}}$.

Two other notable differences are (i) our Definition \ref{def:decay_1} considers uniform bounds with respect to the measure $p$, whereas the decay in \cite[Theorem 43]{chatterji2019online} is with respect to a fixed measure, and (ii) we assume an oblivious adversary, whereas the adversary in \cite[Theorem 43]{chatterji2019online} is non-oblivious.  These differences may potentially be less of a hurdle to overcome compared to the previous paragraph, but they are nevertheless also important.

\subsection{Algorithm and Upper Bound from \cite{iwazaki2026nearly}}
\label{app:discussion_2}

Here we highlight a few differences between our main results and the algorithm and its upper bound from the current work \cite{iwazaki2026nearly}, but note that all of these differences are ultimately minor:
\begin{enumerate}[leftmargin=5ex]
    \item We study the expected regret under an oblivious adversary, whereas \cite{iwazaki2026nearly} studies the pseudoregret under an adaptive adversary.  However, this distinction ultimately has almost no impact on the analysis, and it is straightforward to adapt either analysis to the other setting.
    \item We focus more on the effective dimension $d_*$, whereas \cite{iwazaki2026nearly} focuses more on the maximum information gain $\gamma_T$, but the two are well known to be closely related, e.g., see \cite{calandriello2019gaussian} as well as our \pref{lem:effdim-vs-infogain}.
    \item We use a G-optimal design for our exploration term in \pref{alg:exp2}, whereas \cite{iwazaki2026nearly} uses maximum variance reduction (MVR).  These are essentially equivalent for statistical/regret purposes (though possibly different in terms of computation), since MVR is a greedy Frank-Wolfe like procedure for the log-det (D-optimal design) objective, whose optimality conditions are equivalent to G-optimal design.  In fact, it is explicitly stated in \cite[Appendix I]{iwazaki2026nearly} that exploration via a G-optimal design yields the same regret guarantee.

\end{enumerate}

\end{document}